\documentclass{article}

\usepackage{PRIMEarxiv}

\usepackage[utf8]{inputenc} 
\usepackage[T1]{fontenc}    
\usepackage{hyperref}       
\usepackage{url}            
\usepackage{booktabs}       
\usepackage{amsfonts}       
\usepackage{nicefrac}       
\usepackage{microtype}      
\usepackage{lipsum}
\usepackage{fancyhdr}       
\usepackage{graphicx}       
\graphicspath{{media/}}     

\usepackage{amssymb}
\usepackage{amsmath}
\usepackage{multirow}

\title{HA-ViD: A Human Assembly Video Dataset for Comprehensive Assembly Knowledge Understanding}

\author{
 Hao Zheng \thanks{ These authors contributed equally to this work.} \\
  \texttt{hzhe951@aucklanduni.ac.nz} \\
   \And
 Regina Lee $^*$ \\
  \texttt{klee702@aucklanduni.ac.nz} \\
  \And
 Yuqian Lu \thanks{Corresponding author.}\\
  Department of Mechanical and Mechatronics Engineering\\
  The University of Auckland\\
  \texttt{yuqian.lu@auckland.ac.nz} \\
}

\begin{document}
\maketitle

\begin{abstract}
Understanding comprehensive assembly knowledge from videos is critical for futuristic ultra-intelligent industry. To enable technological breakthrough, we present HA-ViD – the first human assembly video dataset that features representative industrial assembly scenarios, natural procedural knowledge acquisition process, and consistent human-robot shared annotations. Specifically, HA-ViD captures diverse collaboration patterns of real-world assembly, natural human behaviors and learning progression during assembly, and granulate action annotations to subject, action verb, manipulated object, target object, and tool. We provide 3222 multi-view, multi-modality videos (each video contains one assembly task), 1.5M frames, 96K temporal labels and 2M spatial labels. We benchmark four foundational video understanding tasks: action recognition, action segmentation, object detection and multi-object tracking. Importantly, we analyze their performance for comprehending knowledge in assembly progress, process efficiency, task collaboration, skill parameters and human intention. Details of HA-ViD is available at: \url{https://iai-hrc.github.io/ha-vid}
\end{abstract}


\setcounter{footnote}{0}
\section{Introduction}
\label{section1}
Assembly knowledge understanding from videos is crucial for futuristic ultra-intelligent industrial applications, such as robot skill learning \cite{Duque2019}, human-robot collaborative assembly \cite{Lamon2019} and quality assurance \cite{Frustaci2020}. To enable assembly video understanding, a video dataset is required. Such a video dataset should (1) represent real-world assembly scenarios and (2) capture the comprehensive assembly knowledge via (3) a consistent annotation protocol that aligns with human and robot assembly comprehension. However, existing datasets cannot meet these requirements.

First, the assembled products in existing datasets are either too scene-specific \cite{Cicirelli2022,Ben-Shabat2021,Sener2022,Toyer2017,Zhang2020,Ragusa2021} or lack typical assembly parts and tools \cite{Ben-Shabat2021,Sener2022,Toyer2017,Ragusa2021}. Second, existing datasets did not design assembly tasks to foster the emergence of natural behaviors (e.g., varying efficiency, alternative routes, pauses and errors) during procedural knowledge acquisition. Third, thorough understanding of nuanced assembly knowledge is not possible via existing datasets as they fail to annotate subjects, objects, tools and their interactions in a systematic approach.

\begin{figure}[h!]
  \centering
  \includegraphics[width=\linewidth]{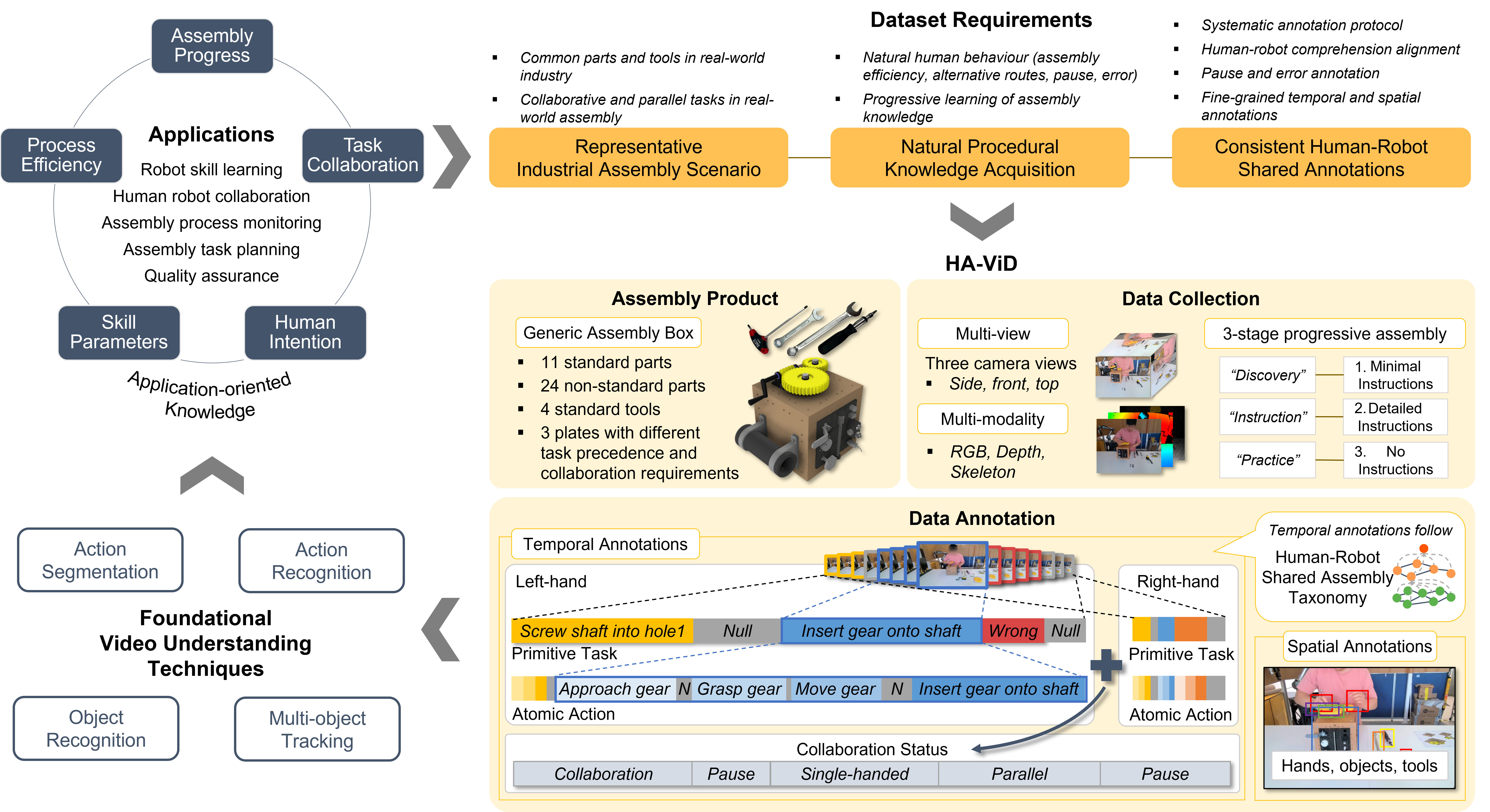}
  \caption{HA-ViD, a dataset designed for industrial applications, represents real-world assembly scenarios, and captures the process of acquiring procedural knowledge. The consistent annotation follows a human-robot shared taxonomy. The dataset features 3222 multi-view, multi-modalities videos (each video contains one task), 1.5M frames, 96K temporal labels and 2M spatial labels.}
  \label{fig1}
\end{figure}

Therefore, we introduce HA-ViD: a human assembly video dataset recording people assembling the Generic Assembly Box (GAB, see Figure \ref{fig1}). We benchmark on four foundational tasks: action recognition, action segmentation, object detection and multi-object tracking (MOT), and analyze their performance for comprehending application-oriented knowledge. HA-ViD features three novel aspects:

\begin{itemize}

\item \textbf{Representative industrial assembly scenarios}: GAB includes 35 standard and non-standard parts frequently used in real-world industrial assembly scenarios and requires 4 standard tools to assemble it. The assembly tasks are arranged onto 3 plates featuring different task precedence and collaboration requirements to promote the emergence of two-handed collaboration and parallel tasks. Different from existing assembly video datasets, GAB represents generic industrial assembly scenarios (see Table \ref{table1}).
\item \textbf{Natural procedural knowledge acquisition process}: Progressive observation, thought and practice process (shown as varying efficiency, alternative assembly routes, pauses, and errors) in acquiring and applying complex procedural assembly knowledge is captured via the designed three-stage progressive assembly setup (see Figure \ref{fig1}). Such a design allows in-depth understanding of the human cognition process, where existing datasets lack (see Table \ref{table1}).
\item \textbf{Consistent human-robot shared annotations}: We designed a consistent fine-grained hierarchical task/action annotation protocol following a Human-Robot Shared Assembly Taxonomy (HR-SAT\footnote{HR-SAT, developed by the same authors, is a hierarchical assembly task representation schema that both humans and robots can comprehend. See details via: \url{https://iai-hrc.github.io/hr-sat}} , to be introduced in Section 2.3). Using this protocol, we, for the first-time, (1) granulate action annotations to subject, action verb, manipulated object, target object, and tool; (2) provide collaboration status annotations via separating two-handed annotations; and (3) annotate human pauses and errors. Such detailed annotation embeds more knowledge sources for diverse understanding of application-oriented knowledge (see Table \ref{table1}).
\end{itemize}

\begin{table}[h!]
\centering
\caption{Comparison between HA-ViD and other assembly video datasets.}
\label{table1}
\resizebox{\textwidth}{!}{%
\begin{tabular}{cccccclccccccc}
\hline
\multirow{2}{*}{Dataset} & \multirow{2}{*}{\begin{tabular}[c]{@{}c@{}}Assembled \\ product\end{tabular}} & \multicolumn{4}{c}{\begin{tabular}[c]{@{}c@{}}Natural procedual \\ knowledge aquisition process\end{tabular}} &  & \multicolumn{6}{c}{\begin{tabular}[c]{@{}c@{}}Consistent human-robot \\ shared assembly taxonomy\end{tabular}} & \multirow{2}{*}{\begin{tabular}[c]{@{}c@{}}Two-handed \\ collaboration \\ status\end{tabular}} \\ \cline{3-6} \cline{8-13}
 &  & \begin{tabular}[c]{@{}c@{}} Varying assembly \\ efficiency\end{tabular} & \begin{tabular}[c]{@{}c@{}}Alternative \\ route\end{tabular} & Pause & Error &  & Subject & \begin{tabular}[c]{@{}c@{}}Action \\ verb\end{tabular} & \begin{tabular}[c]{@{}c@{}}Manipulated \\ object\end{tabular} & \begin{tabular}[c]{@{}c@{}}Target \\ object\end{tabular} & Tool & Two-hand &  \\ \cline{1-6} \cline{8-14} 
Wooden box \cite{Zhang2020} & Wooden box & × & × & × & × &  & × & \checkmark & × & × & \checkmark & × & × \\
IKEA-FA \cite{Toyer2017} & Furniture & × & \checkmark & \checkmark & × &  & × & \checkmark & \checkmark & × & × & × & × \\
MECCANO \cite{Ragusa2021} & Toy motorbike & × & \checkmark & × & × &  & × & \checkmark & \checkmark & × & \checkmark & × & × \\
IKEA ASM \cite{Ben-Shabat2021} & Furniture & × & \checkmark & \checkmark & × &  & × & \checkmark & \checkmark & × & × & × & × \\
Assembly101 \cite{Sener2022} & Toy cars & × & \checkmark & × & \checkmark &  & × & \checkmark & \checkmark & × & \checkmark & × & × \\
HA4M \cite{Cicirelli2022} & \begin{tabular}[c]{@{}c@{}}Epicyclic \\ Gear Train\end{tabular} & × & \checkmark & \checkmark & × &  & × & \checkmark & \checkmark & × & × & × & × \\
\begin{tabular}[c]{@{}c@{}}HA-ViD \\ (ours)\end{tabular} & \begin{tabular}[c]{@{}c@{}}Generic \\ assembly box\end{tabular} & \checkmark & \checkmark & \checkmark & \checkmark &  & \checkmark & \checkmark & \checkmark & \checkmark & \checkmark & \checkmark & \checkmark \\ \hline
\end{tabular}%
}
\end{table}

\section{Dataset}
\label{section2}

In this section, we present the process of building HA-ViD and provide essential statistics.

\subsection{Generic Assembly Box}

To ensure the dataset can represent real-world industrial assembly scenarios, we designed the GAB shown in Figure \ref{fig1}.

First, GAB\footnote{Find GAB CAD files at: \url{https://iai-hrc.github.io/ha-vid}.} is a 250$\times$250$\times$250mm box including 11 standard and 24 non-standard parts frequently used in real-world industrial assembly. Four standard tools are required for assembling GAB. The box design also allows participants to naturally perform tasks on a top or side-facing plate, closer to the flexible setups of real-world assembly.

\begin{figure}[h!]
  \centering
  \includegraphics[width=\linewidth]{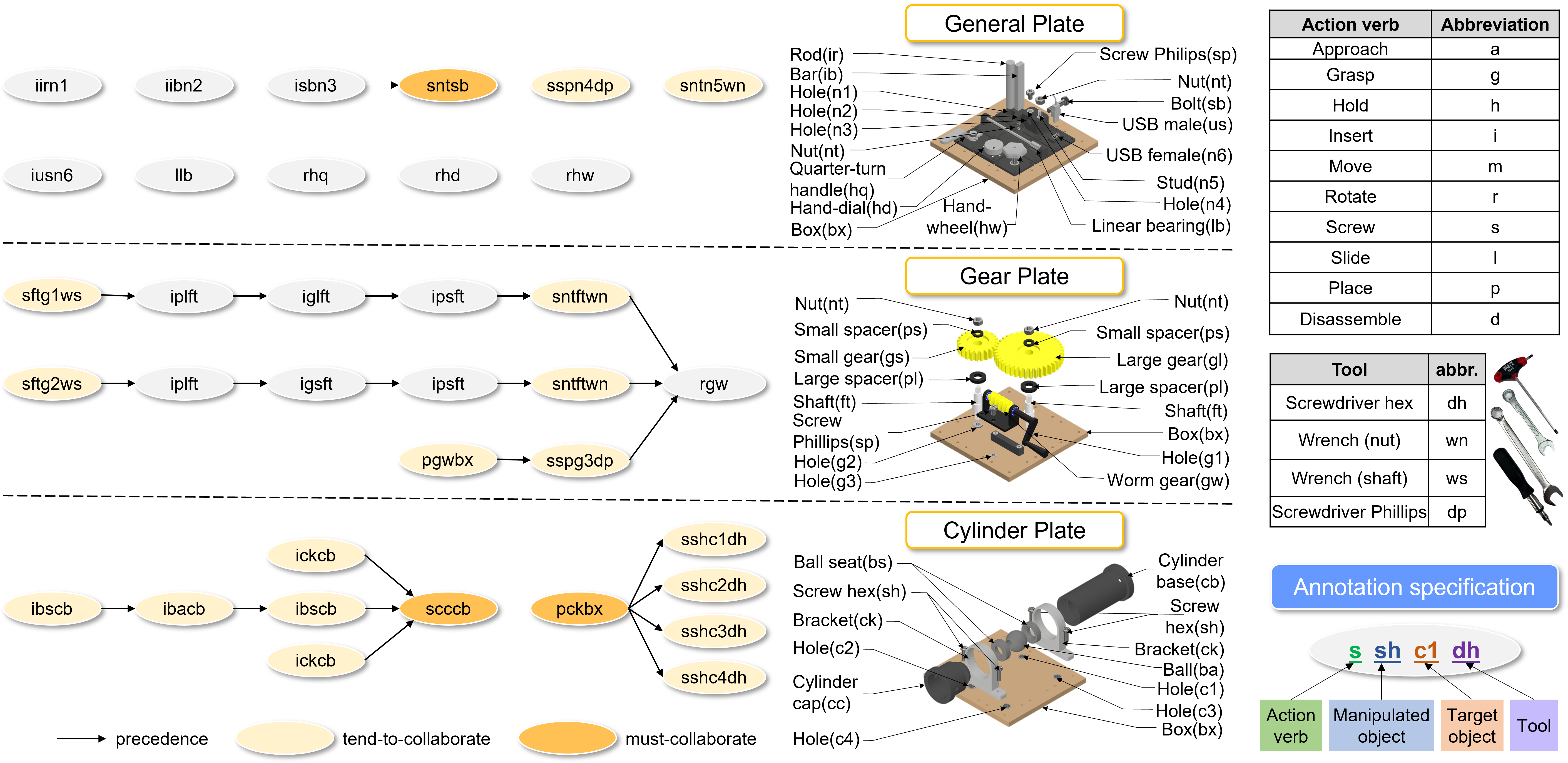}
  \caption{Subject-agnostic task precedence graphs for three plates and annotation specification. “must-collaborate” denotes the task requires two-handed collaboration, and “tend-to-collaborate” denotes the task that tend to need two hands.}
  \label{fig2}
\end{figure}

Second, GAB consists of three plates featuring different task precedence and collaboration requirements. Figure \ref{fig2} shows the subject-agnostic task precedence graphs (SA-TPG) for the three plates with different precedence constraints. These different task precedence graphs provide contextual links between actions, enabling situational action understanding with different complexities. The cylinder plate also has more collaboration tasks, posing greater challenges for understanding collaborative assembly tasks. Gear and cylinder plates contain parts that become hidden after assembly, e.g., spacers under the gears. This introduces additional complexities for understanding assembly status.

\subsubsection{Dataset Collection}
Data was collected on three Azure Kinect RGB+D cameras mounted to an assembly workbench facing the participant from left, front and top views, as shown in Figure \ref{fig4}. Videos were recorded at 1280$\times$720 RGB resolution and 512$\times$512 depth resolution under both lab lighting and natural lighting conditions. 30 participants (15 males, 15 females) assembled each plate 11 to 12 times during a 2-hour session.

To capture the progression of human procedural knowledge \cite{Georgeff1986} acquisition and behaviors (e.g., varying efficiency, alternative routes, pause, and errors) during learning, a three-stage progressive assembly setup is designed. Inspired by discovery learning \cite{Mayer2004}, we design the three stages as\footnote{The instruction files can be found at \url{https://iai-hrc.github.io/ha-vid}. The detailed instructions were written following HR-SAT to align assembly instructions with our annotations.}: \emph{Discovery} – participants are given minimal exploded view instructions of each plate; \emph{Instruction} – participants are given detailed step-by-step instructions of each plate; \emph{Practice} – participants are asked to complete the task without instruction.

The first stage encourages participants to explore assembly knowledge to reach a goal, the second stage provides targeted instruction to deepen participants’ understanding, and the last stage encourages participants to reinforce their learning via practicing. During \emph{Instruction} and \emph{Practice} stages, the participants were asked to perform the assembly with the plate facing upwards and sideways.

\subsubsection{Dataset Annotations}
We provide temporal and spatial annotations to capture rich assembly knowledge shown in Figure \ref{fig1}.

\begin{figure}[h!]
  \centering
  \includegraphics[width=0.9\linewidth]{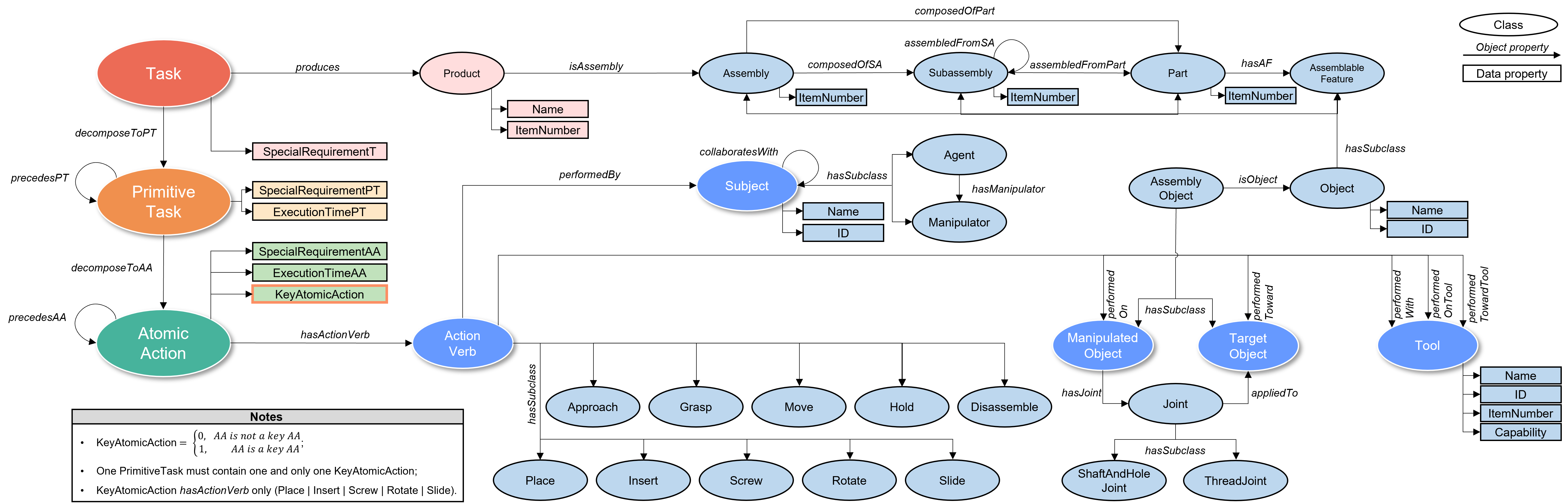}
  \caption[Caption for LOF]%
      {Human-robot shared assembly taxonomy (HR-SAT) schema. We tailored the original taxonomy by removing information that cannot be annotated from videos and incorporating a \emph{Disassemble} action verb to describe human error-and-correction process. We provide textual annotations (see Figure \ref{fig2}) following the typical input formats of current video understanding algorithms. We also offer SA-TPGs as knowledge graphs \footnote in RDF/XML format following the HR-SAT schema to enable advanced assembly knowledge reasoning with enhanced relationship information.}
  \label{fig3}
\end{figure}

\begin{figure}[h!]
  \centering
  \includegraphics[width=\linewidth]{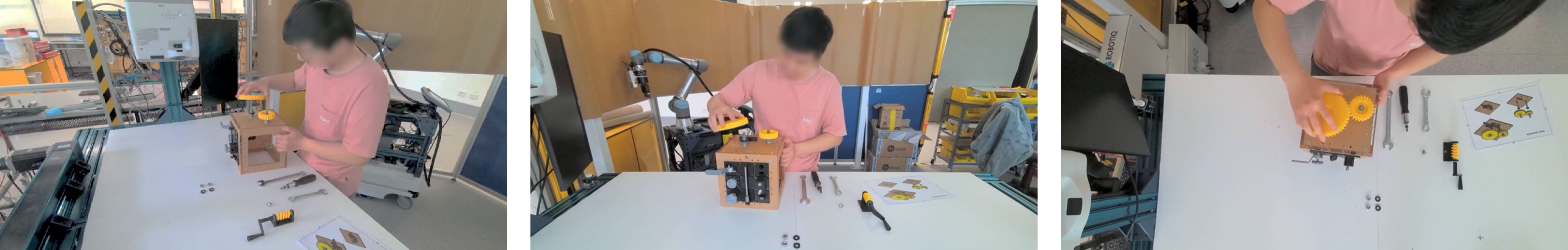}
  \caption{Side, front and top camera views of the workbench.}
  \label{fig4}
  \vspace{-0.2cm}
\end{figure}

To enable human-robot assembly knowledge transfer, the structured temporal annotations are made following HR-SAT. According to HR-SAT (shown in Figure \ref{fig3}), an assembly task can be decomposed into primitive tasks and further into atomic actions. Each primitive task and atomic action contain five description elements: \emph{subject}, \emph{action verb}, \emph{manipulated object}, \emph{target object} and \emph{tool}. Primitive tasks annotations describe a functional change of the manipulated object, such as inserting a gear on a shaft or screwing a nut onto a bolt. Atomic actions describe an interaction change between the subject and manipulated object such as a hand grasping the screw or moving the screw. HR-SAT ensures the annotation transferability, adaptability, and consistency.

\footnotetext{The ST-TPGs files can be downloaded at: \url{https://iai-hrc.github.io/hr-sat}} 

We annotate human pause and error as \emph{null} and \emph{wrong} respectively to enable research on understanding assembly efficiency and learning progression. Our annotations treat each hand as a separate subject. Primitive tasks and atomic actions are labeled for each hand to support multi-subject collaboration related research. Alongside the primitive task annotations, we annotate the two-handed collaboration status as: \emph{collaboration}, when both hand work together on the same task; \emph{parallel}, when each hand is working on a different task; \emph{single-handed}, when only one hand is performing the task while the other hand pauses; and \emph{pause}, when neither hand is performing any task. More details about the temporal annotations can be found in Supplementary Section 2.3.

For spatial annotations, we use CVAT\footnote{\url{https://www.cvat.ai/}}, a video annotation tool, to label bounding boxes for subjects, objects and tools frame-by-frame. Different from general assembly datasets, we treat important assemblable features, such as holes, stud and USB female, as objects, to enable finer-grained assembly knowledge understanding.

\subsection{Statistics}
In total, we collected 3222 videos with side, front and top camera views. Each video contains one task – the process of assembling one plate. Our dataset contains 86.9 hours of footage, totaling over 1.5 million frames with an average of 1 min 37 sec per video (1456 frames). To ensure annotation quality, we manually labeled temporal annotations for 609 plate assembly videos and spatial annotations for over 144K frames. The selected videos for labeling collectively capture the dataset diversity by including videos of different participants, lighting, instructions and camera views.

\begin{figure}[h!]
  \centering
  \vspace{-0.3cm}
  \includegraphics[width=\linewidth]{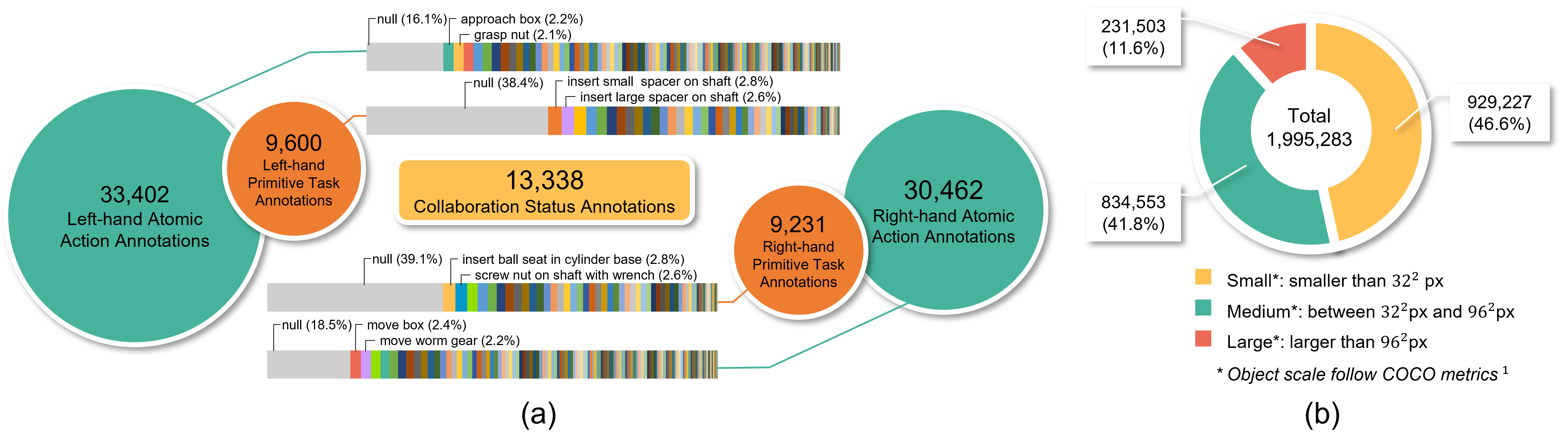}
  \caption{Temporal and spatial annotation statistics. (a) Total number of temporal annotations and annotation distributions, categorized by hands. The three head classes of primitive tasks and atomic actions are shown. (b) Total number of spatial annotations categorized into COCO object scale.}
  \label{fig5}
\end{figure}

Overall, our dataset contains 18831 primitive tasks across 75 classes, 63864 atomic actions across 219 classes, and close to 2M instances of subjects, objects and tools across 42 classes. Figure \ref{fig5} presents the annotation statistics of the dataset. Our dataset shows potential for facilitating small object detection research as 46.6\% of the annotations are of small objects. More statistics can be found in Supplementary Section 2.4. 

\begin{figure}[h!]
  \centering
  \includegraphics[width=\linewidth]{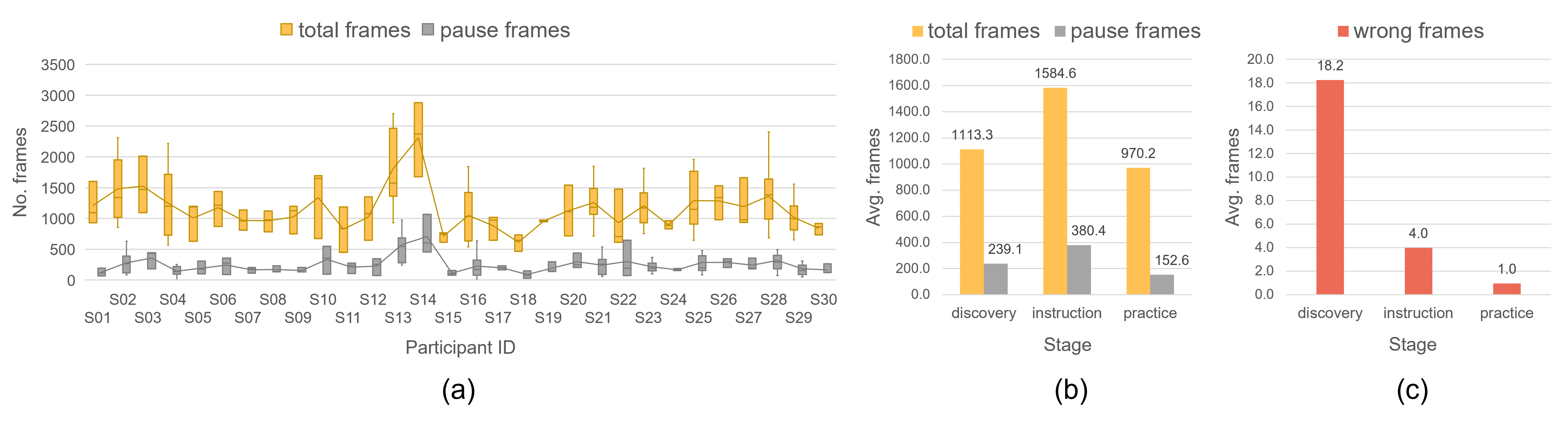}
  \caption{Annotation statistics of total frames, pause frames, and wrong frames. (a) Total frames and pause frames distribution by participant. (b) Average total frames and pause frames per task in each progressive assembly stage. (c) Average wrong frames per task in each progressive assembly stage.}
  \label{fig6}
\end{figure}

Our temporal annotations can be used to understand the learning progression and efficiency of participants over the designed three-stage progressive assembly setup, shown in Figure \ref{fig6}. The combined annotation of \emph{wrong} primitive task, \emph{pause} collaboration status and total frames can indicate features such as errors, observation patterns and task completion time for each participant. Our dataset captures the natural progress of procedural knowledge acquisition, as indicated by the overall reduction in task completion time and pause time from stage 1 to 3, as well as the significant reduction in errors. The \emph{wrong} and \emph{pause} annotations enable research on understanding varying efficiency between participants.

\begin{figure}[h!]
  \centering
  \includegraphics[width=\linewidth]{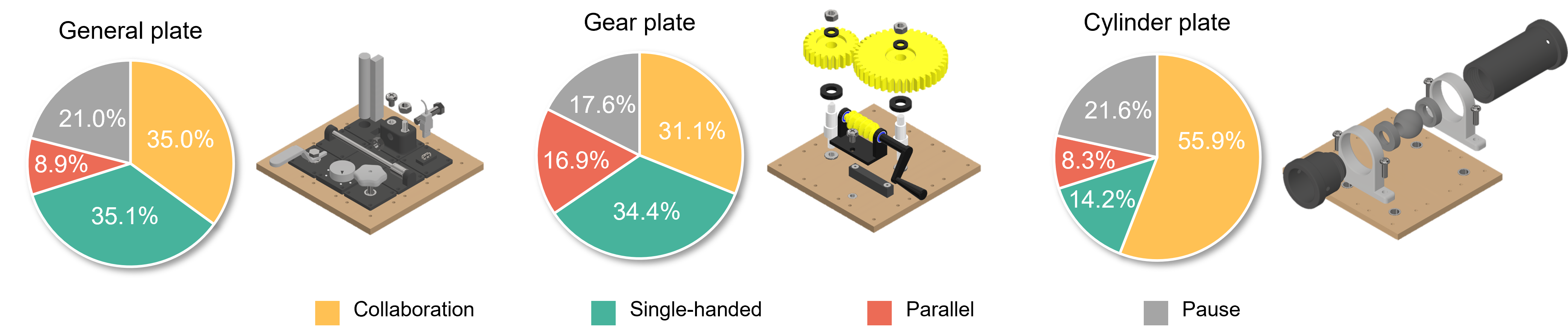}
  \caption{Percentage distribution of each collaboration status annotation for each assembly plate. }
  \label{fig7}
  \vspace{-0.5cm}
\end{figure}

By annotating the collaboration status and designing three assembly plates with different task precedence and collaboration requirements, HA-ViD captures the two-handed collaborative and parallel tasks commonly featured in real-world assembly, shown in Figure \ref{fig7}. Overall, 49.6\% of the annotated frames consist of two-handed tasks. The high percentage of two-handed tasks enables research in understanding the collaboration patterns of complex assembly tasks.

\section{Benchmark Experiments}
\label{section3}
We benchmark SOTA methods for four foundational techniques for assembly knowledge understanding, i.e., action recognition, action segmentation, object detection, and MOT. Due to page limit, we highlight key results and findings in this section, and present implementation details, more results and discussions in the Supplementary Section 3.

\subsection{Action Recognition, Action Segmentation, Object Detection and MOT}

\textbf{Action recognition} is to classify a sequence of video frames into an action category. We split 123 out of 609 temporally labeled videos to be the testset, and the rest is trainset. We benchmark five action recognition methods from three categories: 2D models (TSM \cite{Lin2019}, TimeSFormer \cite{Bertasius2021}), 3D models (I3D \cite{Carreira2017}, MVITv2 \cite{Li2022}), and skeleton-based method (ST-GCN \cite{Yan2018}) and report the Top-1 accuracy and Top-5 accuracy in Table \ref{table2}. 

\textbf{Action segmentation} is to temporally locate and recognize human action segments in untrimmed videos \cite{Wang2021}. Under the same train/test split, we benchmark three action segmentation methods, MS-TCN \cite{Farha2019}, DTGRM \cite{Wang2021} and BCN \cite{Wang2020}, and report the frame-wise accuracy (Acc), segmental edit distance (Edit) and segmental F1 score at overlapping thresholds of 10\% in Table \ref{table3}. 

\textbf{Object detection} is to detect all instances of objects from known classes \cite{Amit2014}. We split 18.4K out of 144K spatially labeled frames to be testset, and the rest is trainset. We benchmark classical two-stage method FasterRCNN \cite{Ren2017}, one-stage method Yolov5 \cite{Jain}, and the SOTA end-to-end Transformer-based method DINO \cite{Zhang2022a} with different backbone networks, and report parameter size (Params), average precision (AP), AP under different IoU thresholds (50\% and 75\%) and AP under different object scales (small, medium and large) in Table \ref{table4}.

\textbf{MOT} aims at locating multiple objects, maintaining their identities, and yielding their individual trajectories given an input video \cite{Luo2021}. We benchmark SORT \cite{Bewley2016} and ByteTrack \cite{Zhang2022} on the detection results of DINO and ground truth annotations (test split of object detection), respectively. We report average multi-object tracking accuracy (MOTA), ID F1 score (IDF1), false positive (FP), false negative (FN), and ID switch (IDS) over the videos in our testing dataset in Table \ref{table5}.

\begin{table}[h!]
\centering
\caption{Baselines of action recognition. Average results over three views are reported here and more detailed results can be found in the Supplementary Section 3.}
\label{table2}
\resizebox{0.9\textwidth}{!}{%
\begin{tabular}{cccclcclcclcc}
\hline
\multirow{3}{*}{Method} & \multirow{3}{*}{View} & \multicolumn{5}{c}{Primitive Task} &  & \multicolumn{5}{c}{Atomic Action} \\ \cline{3-7} \cline{9-13} 
 &  & \multicolumn{2}{c}{Left-Hand} &  & \multicolumn{2}{c}{Right-Hand} &  & \multicolumn{2}{c}{Left-Hand} &  & \multicolumn{2}{c}{Right-Hand} \\ \cline{3-4} \cline{6-7} \cline{9-10} \cline{12-13} 
 &  & Top-1 & Top-5 &  & Top-1 & Top-5 &  & Top-1 & Top-5 &  & Top-1 & Top-5 \\ \cline{1-4} \cline{6-7} \cline{9-10} \cline{12-13} 
ST-GCN \cite{Yan2018} & Average & 39.5 & 60.2 &  & 38.7 & 55.2 &  & 20.3 & 44.4 &  & 19.7 & 40.6 \\
TSM \cite{Lin2019} & Average & 61.0 & \textbf{88.5} &  & 58.6 & \textbf{87.9} &  & 39.6 & 69.4 &  & 37.0 & 67.2 \\
TimeSFormer \cite{Bertasius2021} & Average & 52.1 & 85.4 &  & 51.8 & 84.4 &  & 37.6 & 68.8 &  & 34.6 & 66.1 \\
I3D(rgb+flow) \cite{Carreira2017} & Average & 47.7 & 71.5 &  & 52.9 & 85.1 &  & 43.0 & 75.0 &  & 40.5 & \textbf{72.9} \\
MVITv2 \cite{Li2022} & Average & \textbf{61.5} & 86.3 &  & \textbf{58.7} & 84.1 &  & \textbf{48.4} & \textbf{76.5} &  & \textbf{42.9} & 71.2 \\ \hline
\end{tabular}%
}
\end{table}

\begin{table}[h!]
\centering
\caption{Baselines of action segmentation. Average results over three views are reported here and detailed results can be found in the Supplementary Section 3.}
\label{table3}
\resizebox{0.95\textwidth}{!}{%
\begin{tabular}{ccccclccclccclccc}
\hline
\multirow{3}{*}{Method} & \multirow{3}{*}{View} & \multicolumn{7}{c}{Primitive task} &  & \multicolumn{7}{c}{Atomic Action} \\ \cline{3-9} \cline{11-17} 
 &  & \multicolumn{3}{c}{Left hand} &  & \multicolumn{3}{c}{Right hand} &  & \multicolumn{3}{c}{Left hand} &  & \multicolumn{3}{c}{Right hand} \\ \cline{3-5} \cline{7-9} \cline{11-13} \cline{15-17} 
 &  & F1 & Edit & Acc &  & F1 & Edit & Acc &  & F1 & Edit & Acc &  & F1 & Edit & Acc \\ \cline{1-5} \cline{7-9} \cline{11-13} \cline{15-17} 
MS-TCN \cite{Farha2019} & Avg. & 36.6 & 37.5 & 40.2 &  & 34.7 & 34.8 & 39.3 &  & \textbf{35.1} & 32.5 & \textbf{40.9} &  & \textbf{31.2} & \textbf{32.2} & \textbf{34.6} \\
DTGRM \cite{Wang2021} & Avg. & 39.1 & 37.5 & 40.2 &  & 37.8 & 37.3 & 39.7 &  & 34.3 & \textbf{32.6} & 39.8 &  & 29.8 & 29.3 & 33.1 \\
BCN \cite{Wang2020} & Avg. & \textbf{43.7} & \textbf{41.4} & \textbf{44.1} &  & \textbf{41.3} & \textbf{38} & \textbf{43.4} &  & 18.4 & 15.9 & 39.7 &  & 22.3 & 20.1 & \textbf{34.6} \\ \hline
\end{tabular}%
}
\end{table}

\begin{table}[h!]
\centering
\caption{Baselines of object detection.}
\label{table4}
\resizebox{0.75\textwidth}{!}{%
\begin{tabular}{ccccccccc}
\hline
Method & Backbone & Params & AP & AP50 & AP75 & AP-s & AP-m & AP-l \\ \hline
\multirow{3}{*}{Faster-RCNN \cite{Ren2017}} & ResNet50 & 41.6M & 21.7 & 32.6 & 24.4 & 13.0 & 37.4 & 40.6 \\
 & ResNet101 & 60.6M & 20.9 & 31.1 & 23.9 & 12.3 & \textbf{37.9} & 43.1 \\
 & ResNext101 & 99.5M & 22.2 & 31.6 & 25.7 & 15.0 & 36.2 & 46.2 \\
YOLOv5-s \cite{Jain} & DarkNet & 7.1M & 10.2 & 14.1 & 10.9 & 0.7 & 18.8 & 46.8 \\
YOLOv5-l \cite{Jain} & DarkNet & 46.4M & 12.9 & 17.3 & 14.0 & 1.0 & 28.8 & \textbf{59.8} \\
DINO \cite{Zhang2022a} & Swin-L & \textbf{218M} & \textbf{35.5} & \textbf{54.5} & \textbf{37.7} & \textbf{27.4} & 36.4 & 59.2 \\ \hline
\end{tabular}%
}
\end{table}

\begin{table}[h!]
\centering
\caption{MOT results on object detection results and ground truth object bounding boxes.}
\label{table5}
\resizebox{0.55\textwidth}{!}{%
\begin{tabular}{ccccccc}
\hline
Method & bboxes & MOTA & IDF1 & FP & FN & IDS \\ \hline
\multirow{2}{*}{SORT \cite{Bewley2016}} & dets & \textbf{20.4\%} & 27.1\% & \textbf{737.8} & 9212.3 & \textbf{29} \\
 & gt & 94.5\% & \textbf{69.1\%} & 223.9 & 408.1 & \textbf{54.8} \\ \hline
\multirow{2}{*}{ByteTrack \cite{Zhang2022}} & dets & 20.0\% & \textbf{41.1\%} & 5175.3 & \textbf{4678.3} & 87.2 \\
 & gt & \textbf{98.5\%} & 67.5\% & \textbf{32.4} & \textbf{32.5} & 121.6 \\ \hline
\end{tabular}%
}
\end{table}

The baseline results show that our dataset presents great challenges on the four foundational video understanding tasks compared with other datasets. For example, BCN has 70.4\% accuracy on Breakfast \cite{Kuehne2014}, MViTv2 has 86.1\% Top-1 accuracy on Kinetics-400 \cite{Kay2017}, DINO has 63.3\% AP on COCO test-dev \cite{Lin2014}, and ByteTrack has 77.8\% MOTA on MOT20 \cite{Dendorfer2020}. 

Compared to the above baseline results, we are more concerned with whether existing video understanding methods can effectively comprehend the application-oriented knowledge (in Figure \ref{fig1}). We present our subsequent analysis in Sections 3.2-3.5.

\subsection{Assembly progress}
\textbf{Insight \#1: Assembly action recognition could focus on compositional action recognition and leveraging prior domain knowledge.}
Understanding assembly progress, as an essential application-oriented task, requires real-time action (action verb + interacted objects and tools) recognition, and compare the action history with predefined assembly plan (represented in a task graph). After further analysis of the sub-optimal action recognition performance in Table \ref{table2}, we found recognizing interacting objects and tools are more challenging than recognizing action verbs, (as shown in Table \ref{table6}). Therefore, a promising research direction could be compositional recognizing action verb and interacted objects and tools.  

\begin{table}[h!] 
\centering
\caption{Recall of action verb, manipulated object, target object, and tool recognition, via MVITv2.}
\label{table6}
\resizebox{0.65\textwidth}{!}{%
\begin{tabular}{lllll}
\hline
 & Action verb & Manipulated Object & Target Object & Tool \\ \hline
Primitive Task & 71.1\% & 60.4\% & 57.1\% & 60.8\% \\
Atomic Action & 67.6\% & 50.9\% & 53.5\% & 55.0\% \\ \hline
\end{tabular}%
}
\end{table}

Leveraging prior domain knowledge, such as task precedence and probabilistic correlation between action verbs and feasible objects and tools, one may improve the performance of action recognition. With defined task precedence graphs and rich list of action verb/object/tool pairs, HA-ViD enables research on this aspect.

\textbf{Insight \#2: Assembly action segmentation should focus on addressing under-segmentation issues and improving segment-wise sequence accuracy.} Assembly progress tracking requires obtaining the accurate number of action segments and their sequence. For obtaining the accurate number of action segments from a given video, previous action segmentation algorithms \cite{Wang2021,Farha2019,Wang2020} focused on addressing over-segmentation issues, but lack metrics for quantifying under/over-segmentation. Therefore, we propose segmentation adequacy (SA) to fill this gap. Consider the predicted segments as $s_{\text{pred}}=\{s_{1}',s_{2}',\ldots,s_{F}'\}$ and ground truth segments as $s_{\text{gt}}=\{s_{1},s_{2},\ldots,s_{N}\}$ for a given video, where $F$ and $N$ are the number of segments, $\text{SA} = \tanh\left(\frac{2(F-N)}{F+N}\right)$. Table \ref{table7} reveals the significant under-segmentation issues on our dataset. This reminds the community to pay attention to addressing under-segmentation issues for assembly action understanding. The proposed SA can offer evaluation support, and even assist in designing the loss function as it utilizes hyperbolic tangent function.

\begin{table}[h!]
\centering
\caption{Comparison between our dataset and others on segmentation adequacy. We calculated the average ground truth segment number ($N$), predicted segment number ($F$), and segment adequacy ($SA$) over the videos in the testing datasets of ours and others. The predicted results are from BCN.}
\label{table7}
\resizebox{0.4\textwidth}{!}{%
\begin{tabular}{lllll}
\hline
\multicolumn{2}{l}{Dataset} & $N$ & $F$ & $SA$ \\ \hline
\multirow{2}{*}{HA-ViD(ours)} & Primitive task & 14.9 & 8.3 & -0.47 \\
 & Atomic action & 51.2 & 11.5 & -0.82 \\ \cline{1-2}
\multicolumn{2}{l}{Breakfast} & 6 & 6.8 & -0.12 \\
\multicolumn{2}{l}{GTEA} & 32.5 & 32.9 & -0.03 \\ \hline
\end{tabular}%
}
\end{table}

As for segment-wise sequence accuracy, the low value of Edit in Table \ref{table3} suggests pressing required research efforts. Compared with Breakfast \cite{Kuehne2014} (66.2\% Edit score with BCN algorithm), our dataset presents greater challenges.

\subsection{Process Efficiency}
Understanding process efficiency is essential for real-world industry. It requires video understanding methods to be capable of recognizing human pause and error. HA-ViD supports this research by providing \emph{null} and \emph{wrong} labels.

\textbf{Insight \#3: For \emph{null} action understanding, efforts need to be made on addressing imbalanced class distribution.} Table \ref{table8} shows the recall and precision of action recognition and action segmentation of \emph{null} actions. We suspect the high recall and low precision is caused by the imbalanced class distribution, as null is the largest head class (see Figure \ref{fig5}).

\begin{table}[h!]
\centering
\caption{ Recall and precision of \emph{null} recognition and segmentation. Action recognition results are from MVITv2 and action segmentation results are from BCN.}
\label{table8}
\resizebox{0.4\textwidth}{!}{%
\begin{tabular}{llll}
\hline
\multicolumn{2}{l}{} & Recall & Precision \\ \hline
\multirow{2}{*}{Recognition} & Primitive Task & 90.8\% & 65.1\% \\
 & Atomic Action & 81.5\% & 39.1\% \\
\multirow{2}{*}{Segmentation} & Primitive Task & 80.9 & 45.1\% \\
 & Atomic Action & 84.6\% & 37.5\% \\ \hline
\end{tabular}%
}
\end{table}

\textbf{Insight \#4: New research from \emph{wrong} action annotations.} \emph{Wrong} action is the assembly action (primitive task level) occurred at wrong position or order. Our annotation for \emph{wrong} actions allows in-depth research on understanding its appearing patterns between participants across the three stages. Joint understanding between \emph{wrong} actions and their adjacent actions could also trigger new research of predicting \emph{wrong} actions based on action history. 

\subsection{Task Collaboration}
\textbf{Insight \#5: New research on understanding parallel tasks from both hands} Table \ref{table9} shows that both action recognition and segmentation have lowest performance on parallel tasks during assembly. One possible reason is that the foundational video understanding methods rely on global features of each image, and do not explicitly detect and track the action of each hand. This calls for new methods that can independently track both hands and recognize their actions through local features. Recent research on human-object interaction detection in videos \cite{Tu2022,Chiou2021} could offer valuable insights.

\begin{table}[h!]
\centering
\caption{Recall of two-handed primitive task recognition and segmentation in four collaboration status. Action recognition results are from MVITv2 and action segmentation results are from BCN.}
\label{table9}
\resizebox{0.95\textwidth}{!}{%
\begin{tabular}{cccccclccc}
\hline
\multirow{2}{*}{} & \multicolumn{4}{c}{Action recognition results} & \multicolumn{5}{c}{Action segmentation results} \\ \cline{2-10} 
 & Collaboration & Parallel & Single-handed & Pause & \multicolumn{2}{c}{Collaboration} & Parallel & Single-handed & Pause \\ \hline
Left hand & 52.5\% & 39.7\% & 54.2\% & 92.4\% & \multicolumn{2}{c}{32.1\%} & 15.4\% & 18.5\% & 85.5\% \\
Right hand & 46.1\% & 30.5\% & 50.7\% & 93.3\% & \multicolumn{2}{c}{35.0\%} & 24.2\% & 17.2\% & 82.9\% \\ \hline
\end{tabular}%
}
\end{table}

\subsection{Skill Parameters and Human Intention}
Understanding skill parameters and human intentions from videos is essential for robot skill learning and human-robot collaboration (HRC) \cite{Mees2020,Zheng2022}. 

Typically, skill parameters vary depending on the specific application. However, there are certain skill parameters that are commonly used, including trajectory, object pose, force and torque \cite{Jeon2022,Berger2016}. While videos cannot capture force and torque directly, our dataset offers spatial annotations that enable tracking the trajectory of each object. Additionally, the object pose can be inferred from our dataset via pose estimation methods. Therefore, HA-ViD can support research in this direction.

Understanding human intention in HRC refers to a combination of trajectory prediction, action prediction and task goal understanding \cite{Lu2022}. Our spatial annotations provide trajectory information, SA-TPGs present action sequence constraints, and GAB CAD files offer the final task goals. Therefore, HA-ViD can enhance the research in this aspect.

\section{Conclusion}
We present HA-ViD, a human assembly video dataset, to advance comprehensive assembly knowledge understanding toward real-world industrial applications. We designed a generic assembly box to represent industrial assembly scenarios and a three-stage progressive learning setup to capture the natural process of human procedural knowledge acquisition. The dataset annotation follows a human-robot shared assembly taxonomy. HA-ViD includes (1) multi-view, multi-modality data, fine-grained action annotations (subject, action verb, manipulated object, target object, and tool), (2) human pause and error annotations, and (3) collaboration status annotations to enable technological breakthroughs in both foundational video understanding techniques and industrial application-oriented knowledge comprehension.

As for limitation of HA-ViD, the imbalanced class distribution of primitive tasks and atomic actions could cause biased model performance and insufficient learning. In addition, the true complexities and diversities of real-world assembly scenarios may still not be fully captured.

We benchmarked strong baseline methods of action recognition, action segmentation, object detection and multi-object tracking, and analyzed their performance on comprehending application-oriented knowledge in assembly progress, process efficiency, task collaboration, skill parameter and human intention. The results show that our dataset captures essential challenges for foundational video understanding tasks, and new methods need to be explored for application-oriented knowledge comprehension. We envision HA-ViD will open opportunities for advancing video understanding techniques to enable futuristic ultra-intelligent industry.

\section{Acknowledgements}
This work was supported by The University of Auckland FRDF New Staff Research Fund (No. 3720540).


\newpage
\begin{center}
\Large\textbf{\emph{Supplementary Document} for HA-ViD: A Human Assembly Video Dataset for Comprehensive Assembly Knowledge Understanding}\\
\end{center}

\setcounter{section}{0}
\setcounter{figure}{0}
\setcounter{table}{0}
\setcounter{footnote}{0}
\section{Overview}
This supplementary document contains additional information about HA-ViD.

Section \ref{sup1} further describes the process of building HA-ViD, including the design of the Generic Assembly Box, data collection, data annotation, and annotation statistics.

Section \ref{sup2} presents the implementation details of our baselines, discusses the experimental results, and provides the licenses of the benchmarked algorithms.

Section \ref{sup3} discusses the bias and societal impact of HA-ViD.

Section \ref{sup4} presents the research ethics for HA-ViD.

\section{HA-ViD Construction}
\label{sup1}

In this section, we further discuss the process of building HA-ViD. First, we introduce the design of the Generic Assembly Box. Second, we describe the three-stage data collection process. Third, we describe data annotation details. Finally, we present critical annotation statistics. 

\subsection{Generic Assembly Box Design}
\begin{figure}[h!]
  \centering
  \includegraphics[width=\linewidth]{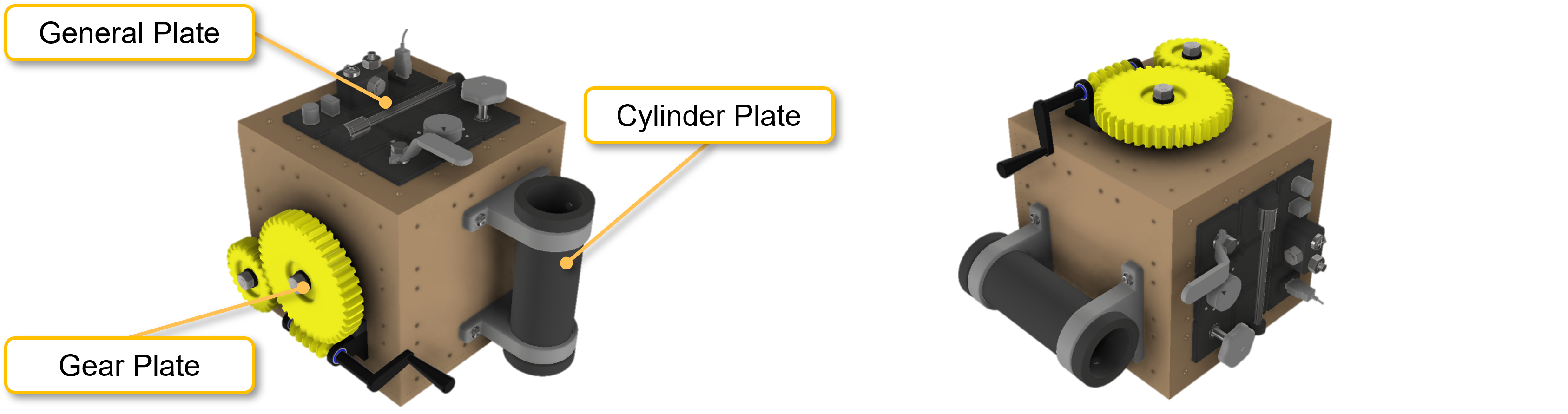}
  \caption{The fully assembled Generic Assembly Box is shown in two different orientations. Each plate can be assembled facing upwards or sideways.}
  \label{figsup1}
\end{figure}

To ensure the dataset is representative of real-world industrial assembly scenarios, we designed the Generic Assembly Box (GAB), a 250$\times$250$\times$250mm box (see Figure \ref{figsup1}), which consists of 11 standard parts and 25 non-standard parts and requires 4 standard tools during assembly (see Figure 2).

\begin{figure}[h!]
  \centering
  \includegraphics[width=0.7\linewidth]{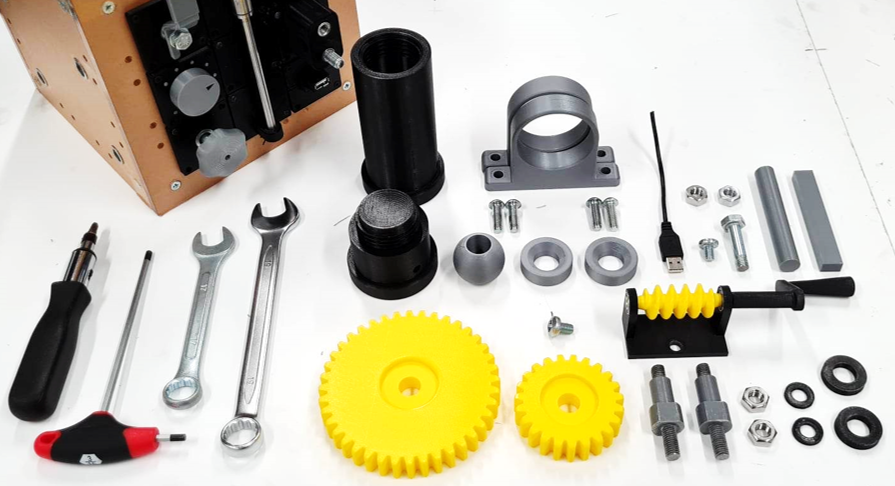}
  \caption{The Generic Assembly Box consists of 11 standard parts and 25 non-standard parts and requires 4 different standard tools during assembly.}
  \label{figsup2}
\end{figure}

GAB has three assembly plates, including \textbf{General Plate}, \textbf{Gear Plate}, and \textbf{Cylinder Plate}, and three blank plates. The opposite face of each assembly plate is intentionally left blank to allow a different assembly orientation. Three assembly plates feature different design purposes.

\begin{figure}[h!]
  \centering
  \includegraphics[width=\linewidth]{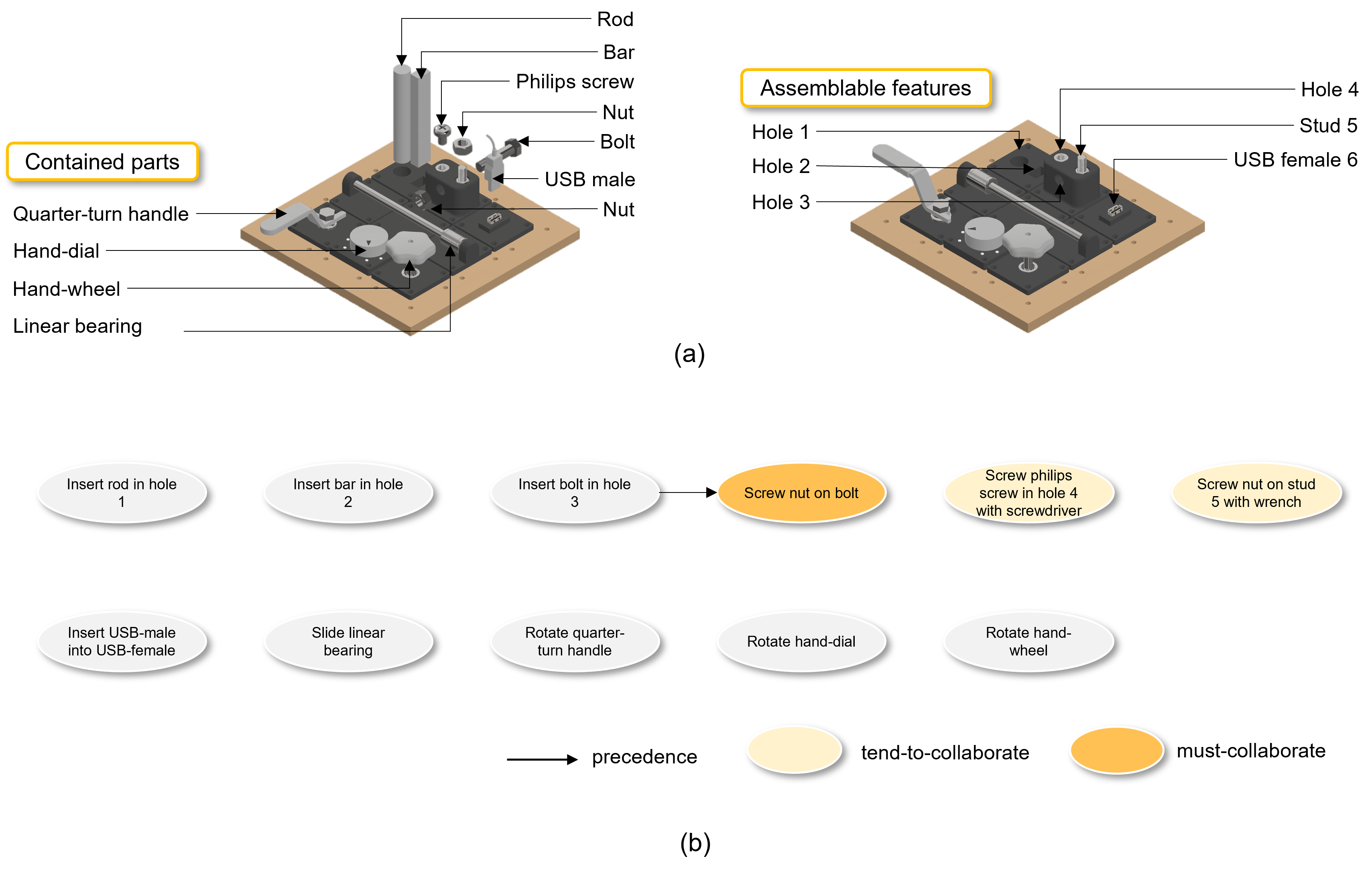}
  \caption{The general plate (a) the contained parts and assemblable features and (b) subject-agnostic task precedence graph where “must-collaborate” denotes the task requires two-handed collaboration, and “tend-to-collaborate” denotes the task that tend to need two hands. Different from general assembly datasets, we treat assemblable features, such as holes, stud and USB female, as objects, to enable finer-grained assembly knowledge understanding.}
  \label{figsup3}
\end{figure}

\textbf{General Plate} (see Figure \ref{figsup3}) was designed to capture action diversity. The general plate consists of 11 different parts. The parts used in this plate were designed to include the different directions, shapes, and forces in which the common assembly actions can be performed. Since there is close to no precedence between assembling different parts, General Plate results in the most variety of possible assembly sequences.

\begin{figure}[h!]
  \centering
  \includegraphics[width=\linewidth]{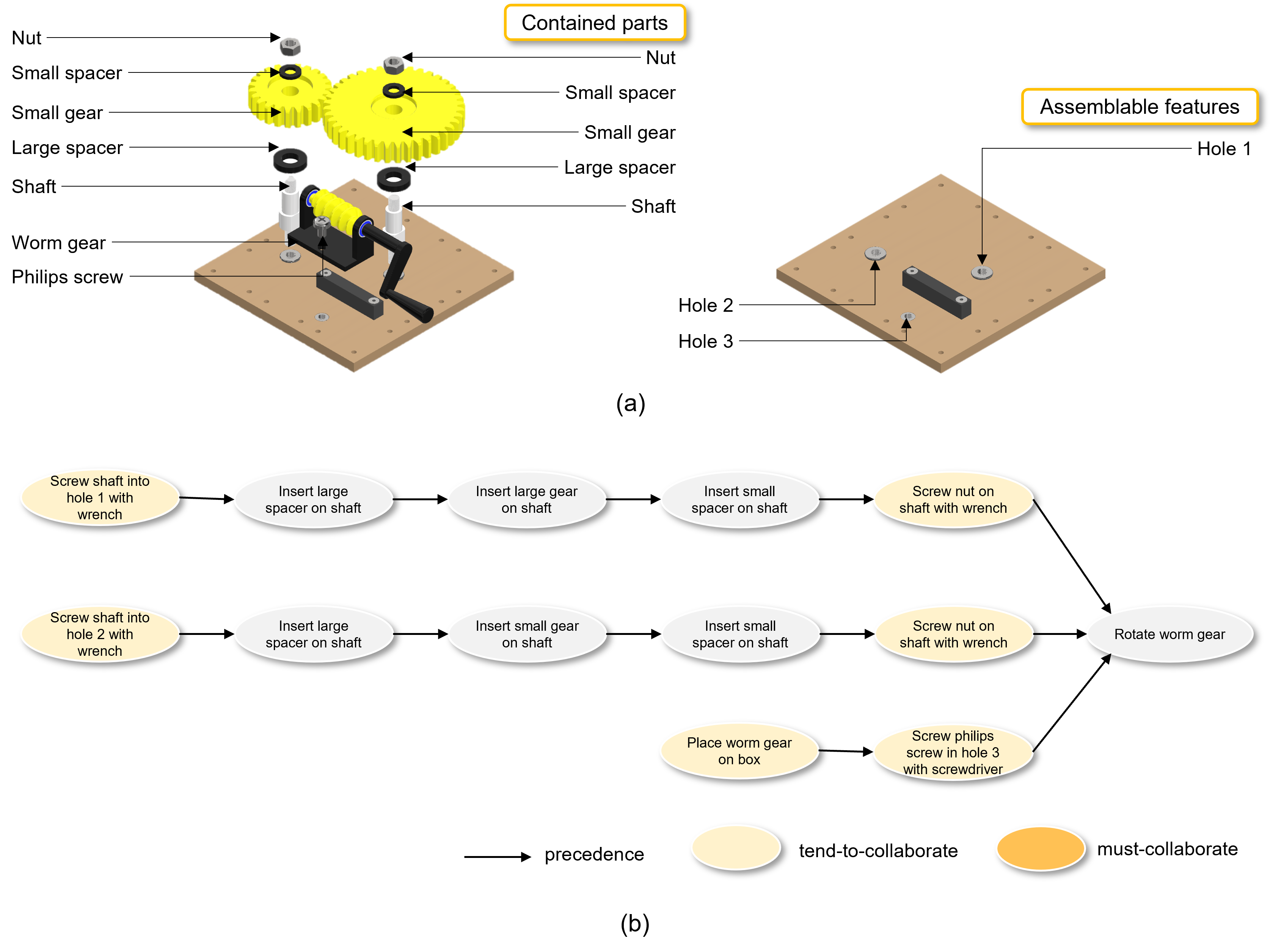}
  \caption{The gear plate (a) the contained parts and assemblable features and (b) subject-agnostic task precedence.}
  \label{figsup4}
\end{figure}

\textbf{Gear Plate} (see Figure \ref{figsup4}) was designed to capture parallel two-handed tasks, e.g., two hands inserting two spur gears at the same time. Gear Plate has three gear sub-systems: large gear, small gear, and worm gear, which mesh together to form a gear mechanism. The plate consists of 12 different parts. Gear Plate has a higher precedence constraint on assembly sequence than the general plate.

\begin{figure}[h!]
  \centering
  \includegraphics[width=\linewidth]{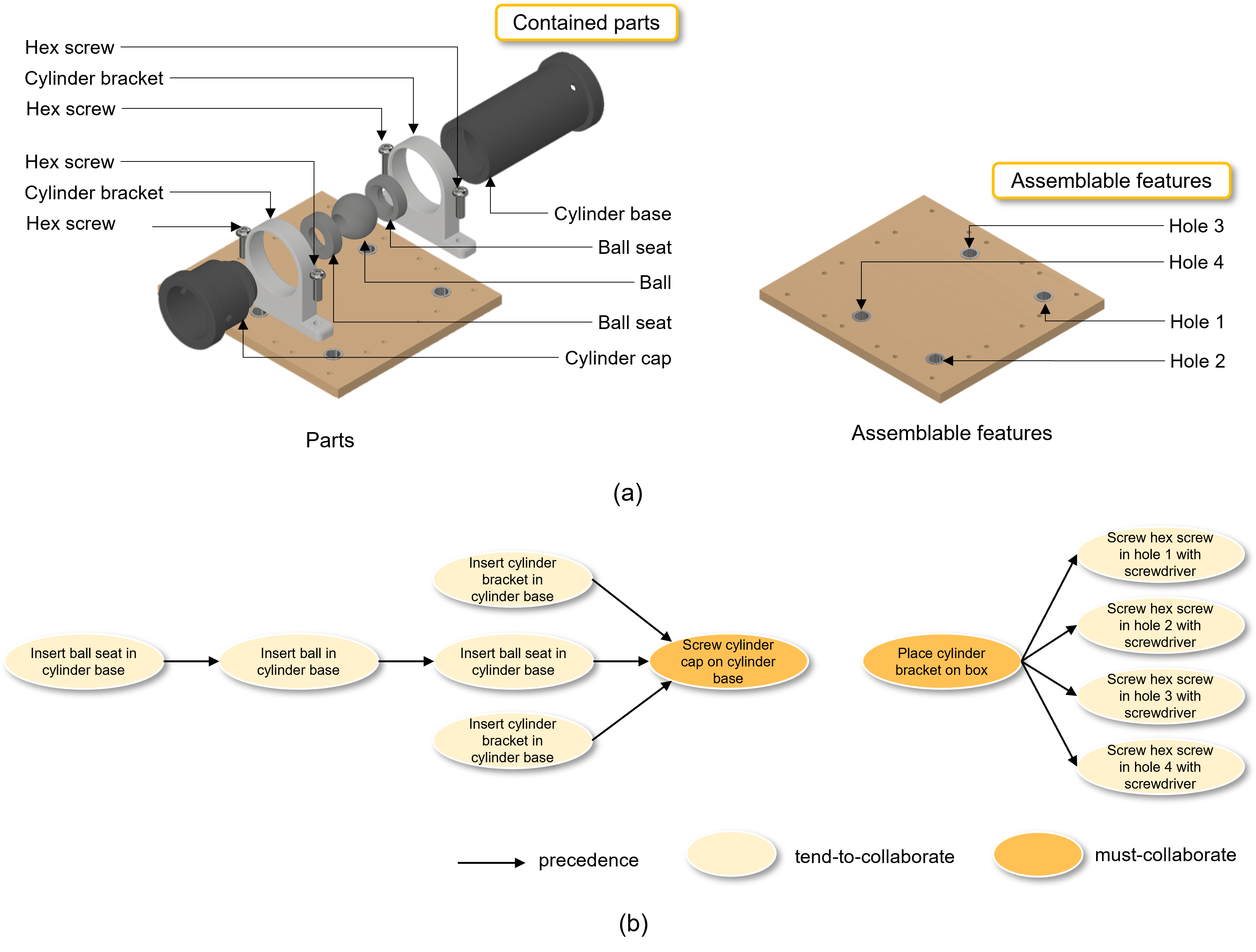}
  \caption{The cylinder plate (a) the contained parts and assemblable features and (b) subject-agnostic task precedence.}
  \label{figsup5}
\end{figure}

\textbf{Cylinder Plate} (see Figure \ref{figsup5}) was designed to capture two-handed collaborative tasks, e.g., two hands collaborating on screwing the cylinder cap onto the cylinder base. Cylinder Plate requires assembling a cylinder subassembly and fastening it onto the plate. This plate consists of 11 parts. The parts were designed to represent assembling a subassembly where parts become fully occluded or partially constrained to another part (see the cylinder in Figure \ref{figsup5}). 

Table \ref{tablesup1} shows a summary of the three assembly plates. The box can be easily replicated using standard components, laser cutting, and 3D printing. The CAD files and bill of material can be downloaded from our website\footnote{\url{https://iai-hrc.github.io/ha-vid}}.

\begin{table}[]
\centering
\caption{Summary of the three Generic Assembly Box plates.}
\label{tablesup1}
\resizebox{\textwidth}{!}{%
\begin{tabular}{ccccccc}
\hline
Plate & Design purpose & \begin{tabular}[c]{@{}c@{}}Precedence \\ constraint\end{tabular} & \begin{tabular}[c]{@{}c@{}}Two-handed \\ collaboration\end{tabular} & \begin{tabular}[c]{@{}c@{}}Standard \\ Parts\end{tabular} & \begin{tabular}[c]{@{}c@{}}Non-standard \\ parts\end{tabular} & Tools \\ \hline
General & \begin{tabular}[c]{@{}c@{}}Action and assembly sequence variety \\ and minimal precedence.\end{tabular} & Minimal & Low & 4 & 7 & 2 \\
Gear & Parallel tasks and high precedence. & High & Medium & 3 & 9 & 3 \\
Cylinder & Collaboration tasks and high precedence. & High & High & 4 & 7 & 1 \\ \hline
\end{tabular}%
}
\end{table}

\subsection{Data Collection}
\begin{figure}[h!]
  \centering
  \includegraphics[width=\linewidth]{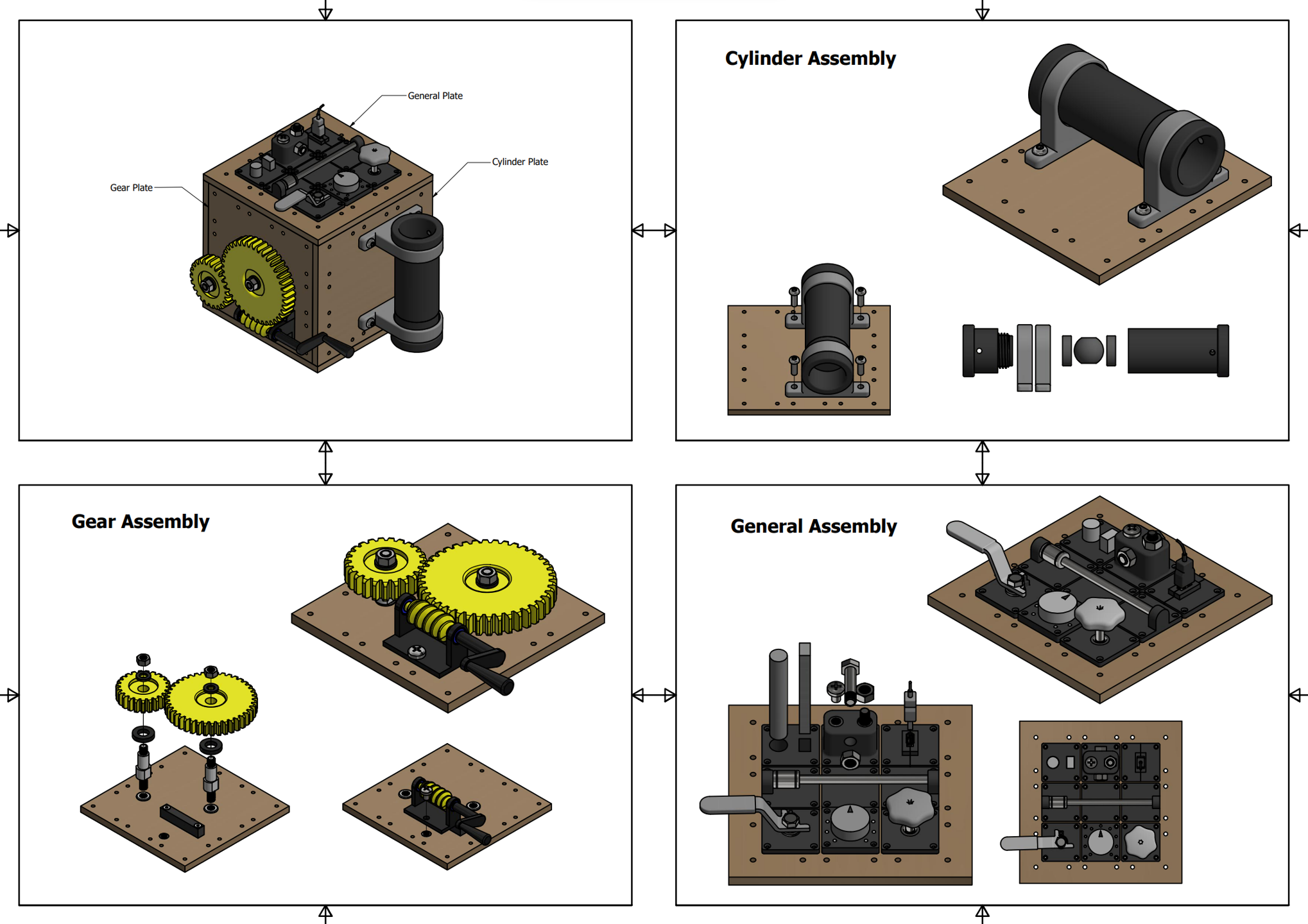}
  \caption{Minimal instruction pages.}
  \label{figsup6}
\end{figure}

Data was collected on three Azure Kinect RGB+D cameras mounted to an assembly workbench. 30 participants (15 male, 15 female) were recruited for a 2-hour session to assemble the GAB. During the data collection session, participants were given a fully disassembled assembly box, assembly parts, tools, and instructions. To capture the natural progress of human procedural knowledge acquisition and behaviors (varying efficiency, alternative routes, pauses, and errors), we designed a three-stage progressive assembly setup:

\emph{\textbf{Discovery}}: Participants were asked to assemble a plate twice following the minimal visual instructions (see Figure \ref{figsup6}).

\begin{figure}[h!]
  \centering
  \includegraphics[width=\linewidth]{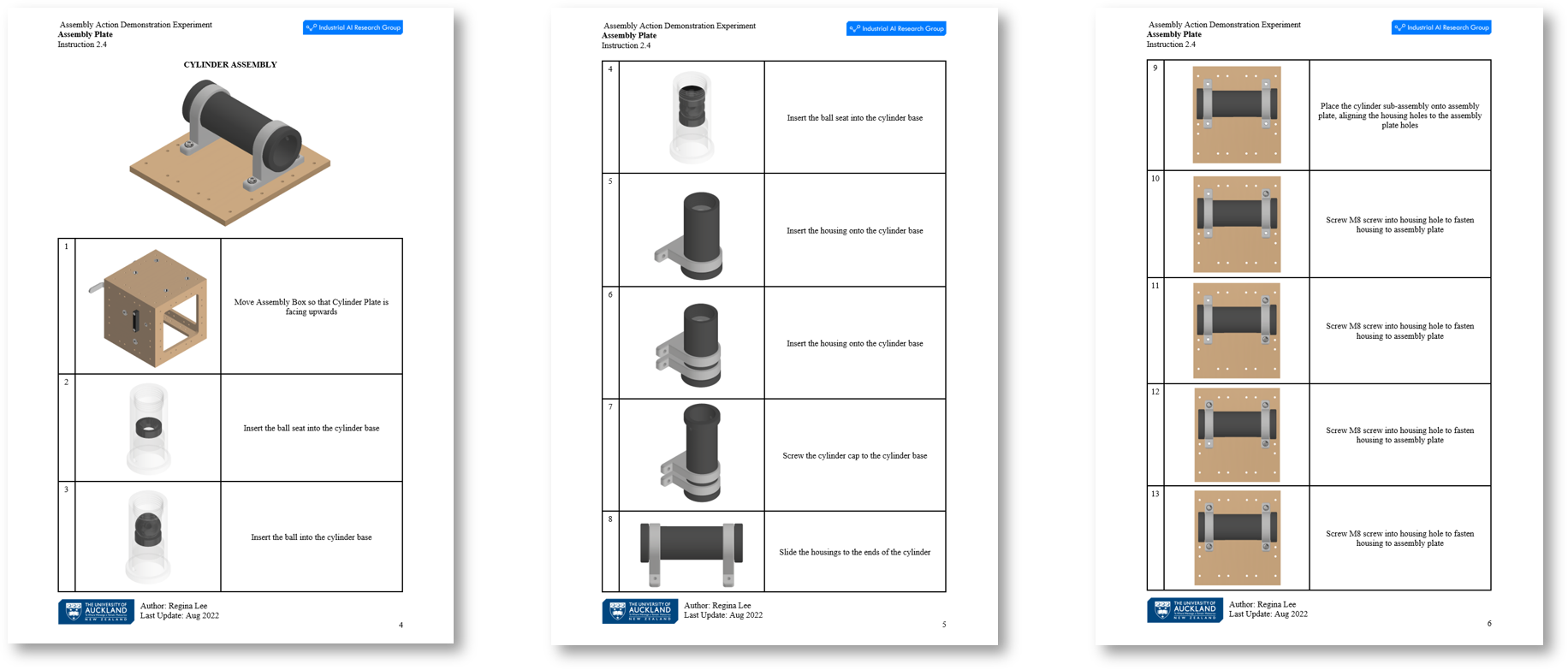}
  \caption{Example of the detailed instruction provided to participants for the cylinder assembly plate.}
  \label{figsup7}
\end{figure}

\emph{\textbf{Instruction}}: Participants were asked to assemble a plate six times following the detailed step-by-step instructions (see Figure \ref{figsup7}). Six different instruction versions were created, each presenting a different assembly sequence. Each participant was given three different instruction versions, where two attempts were completed following each instruction version. The three instruction versions given to one participant must contain assembling the plate facing both upwards and sideways. 

\emph{\textbf{Practice}}: After the first two stages, participants were asked to assemble a plate four times without any instructions. During this stage, participants performed two attempts of each plate facing upwards and two attempts of each plate facing sideways.

The instruction files are available on our website\footnote{https://iai-hrc.github.io/ha-vid}.

\subsection{Data Annotation}
\label{sup2.3}
To capture rich assembly knowledge, we provide temporal and spatial annotations. 

\begin{figure}[h!]
  \centering
  \includegraphics[width=\linewidth]{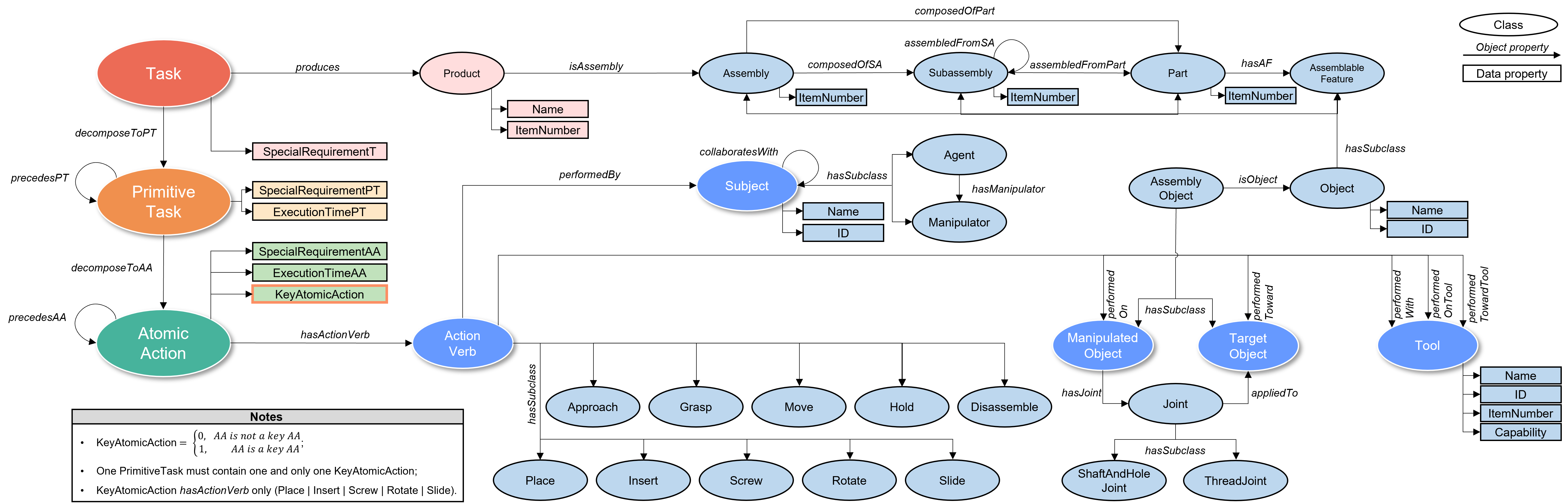}
  \caption{Human-robot shared assembly taxonomy (HR-SAT) schema. We tailored the original taxonomy by removing information that cannot be annotated from videos and incorporating a \emph{Disassemble} action verb to describe human error-and-correction process.}
  \label{figsup8}
\end{figure}

\begin{figure}[h!]
  \centering
  \includegraphics[width=\linewidth]{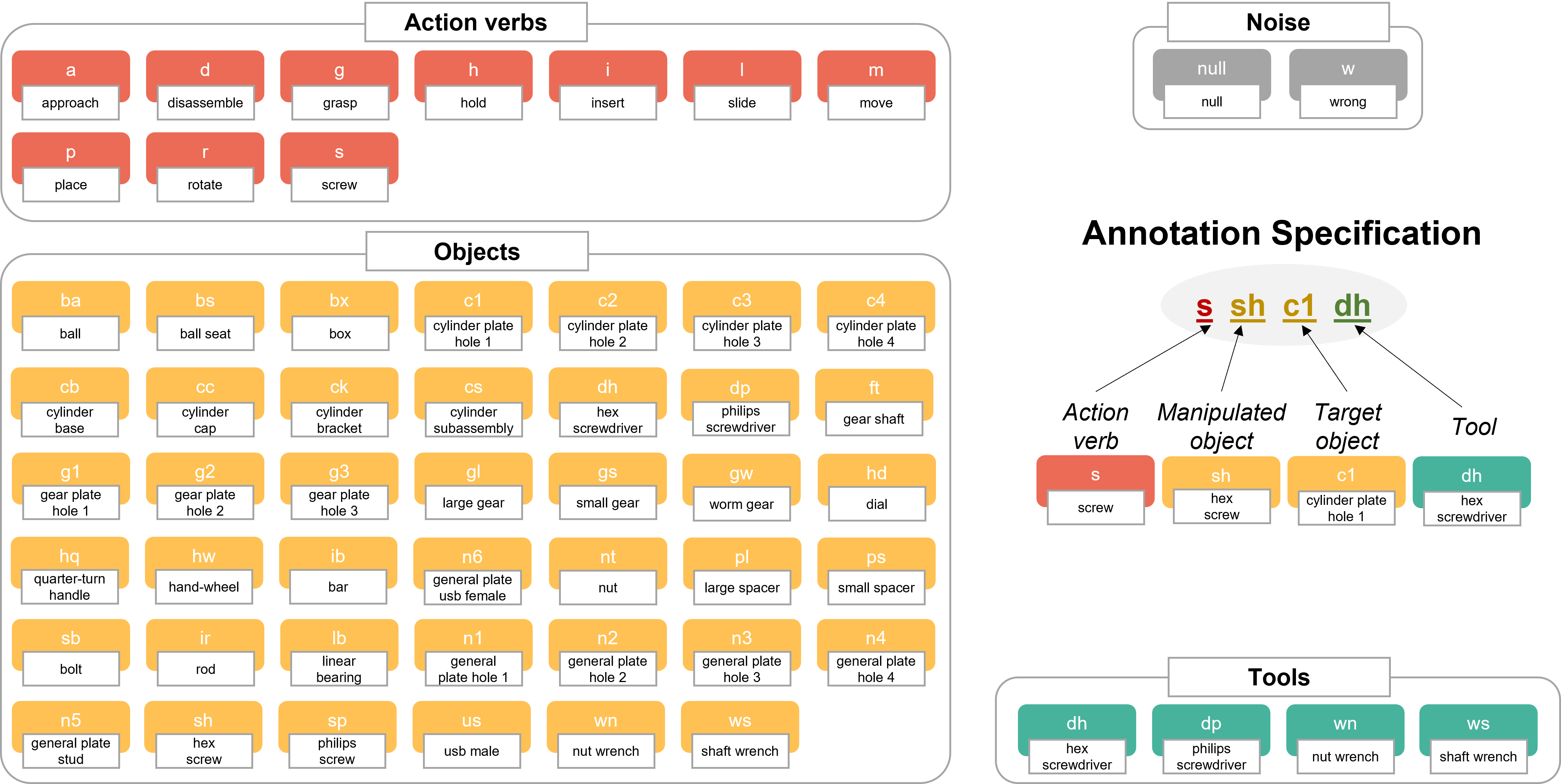}
  \caption{The annotation specification and the list of abbreviated action verbs, objects, and tools annotated in HA-VID.}
  \label{figsup9}
\end{figure}

\textbf{Temporal Annotations}: In HR-SAT\footnote{Details for the definitions of primitive task and atomic action can be found at: https://iai-hrc.github.io/hr-sat}, an assembly task can be decomposed into a series of primitive tasks, and each primitive task can be further decomposed into a series of atomic actions. For both primitive task and atomic action, there are five fundamental description elements: \emph{subject}, \emph{action verb}, \emph{manipulated object}, \emph{target object}, and \emph{tool} (see Figure \ref{figsup8}). We follow HR-SAT to provide primitive task and atomic action annotations for the assembly processes recorded in the videos. To enable the research in two-handed collaboration task understanding, we defined the two hands of each participant as two separate subjects, and we annotated \emph{action verb}, \emph{manipulated object}, \emph{target object}, and \emph{tool} for each \emph{subject}. For both primitive task and atomic action annotations, we follow the annotation specification shown in Figure \ref{figsup9}.

\begin{figure}[h!]
  \centering
  \includegraphics[width=\linewidth]{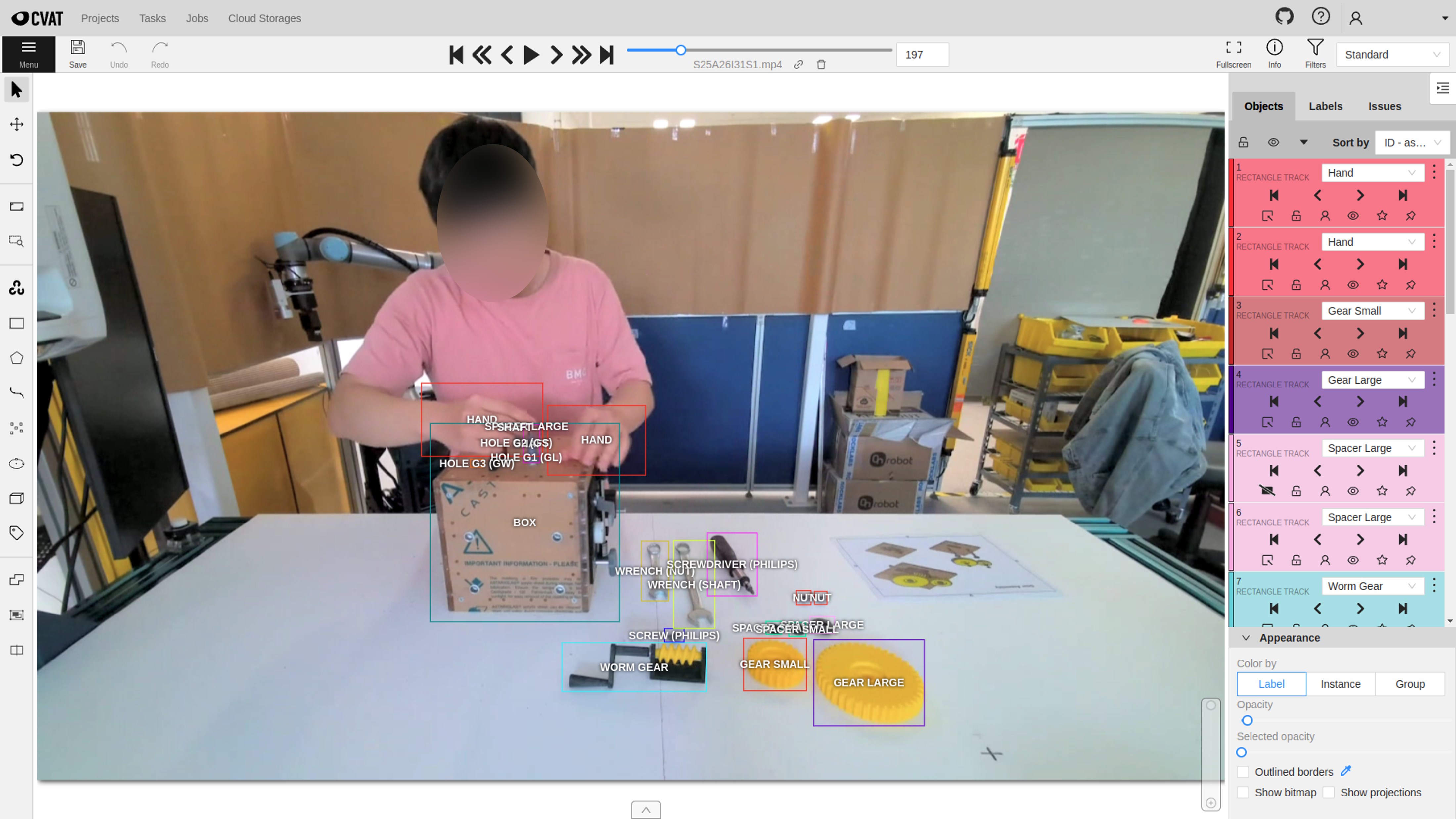}
  \caption{CVAT interface for annotating the subjects (two hands), objects (manipulated object, target object), and tools.}
  \label{figsup10}
\end{figure}

\textbf{Spatial Annotations}: For spatial annotations, we use CVAT\footnote{https://www.cvat.ai/} to annotate the subjects (two hands), objects (manipulated object, target object), and tools via bounding boxes, shown in Figure \ref{figsup10}. 

\subsection{Annotation Statistics}
\label{sup2.4}
\begin{figure}[h!]
  \centering
  \includegraphics[width=\linewidth]{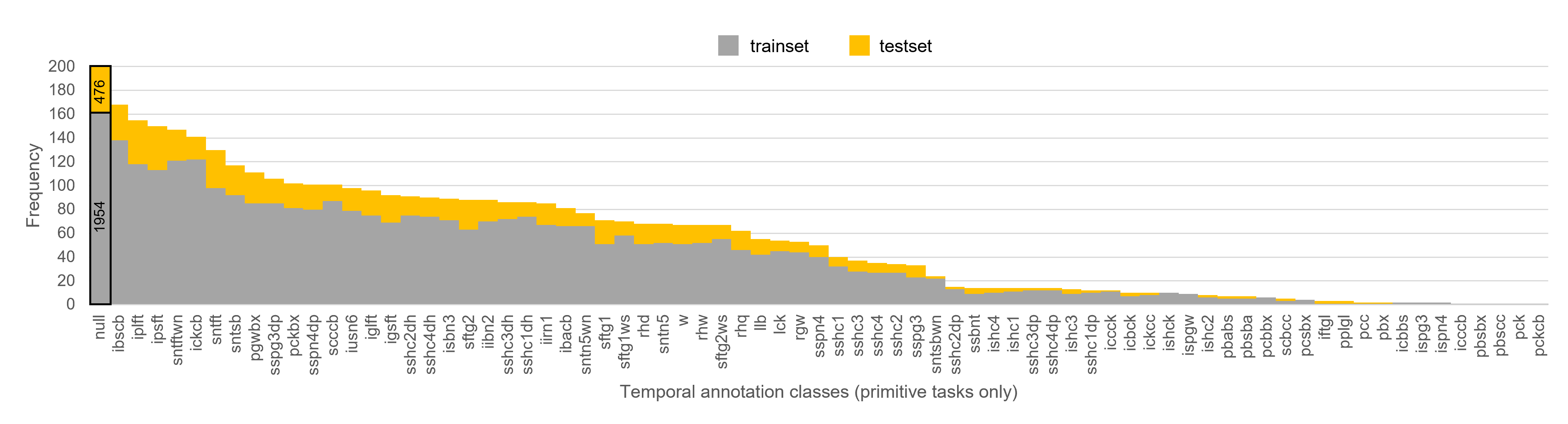}
  \caption{Trainset and testset distribution of the 75 primitive tasks classes. Additionally, to show the distribution better, the frequency axis bound has been reduced, which cuts off the column for the \emph{null} class. Instead, we have manually overwritten the \emph{null} class column with the trainset and testset frequency.}
  \label{figsup11}
\end{figure}

\begin{figure}[h!]
  \centering
  \includegraphics[width=\linewidth]{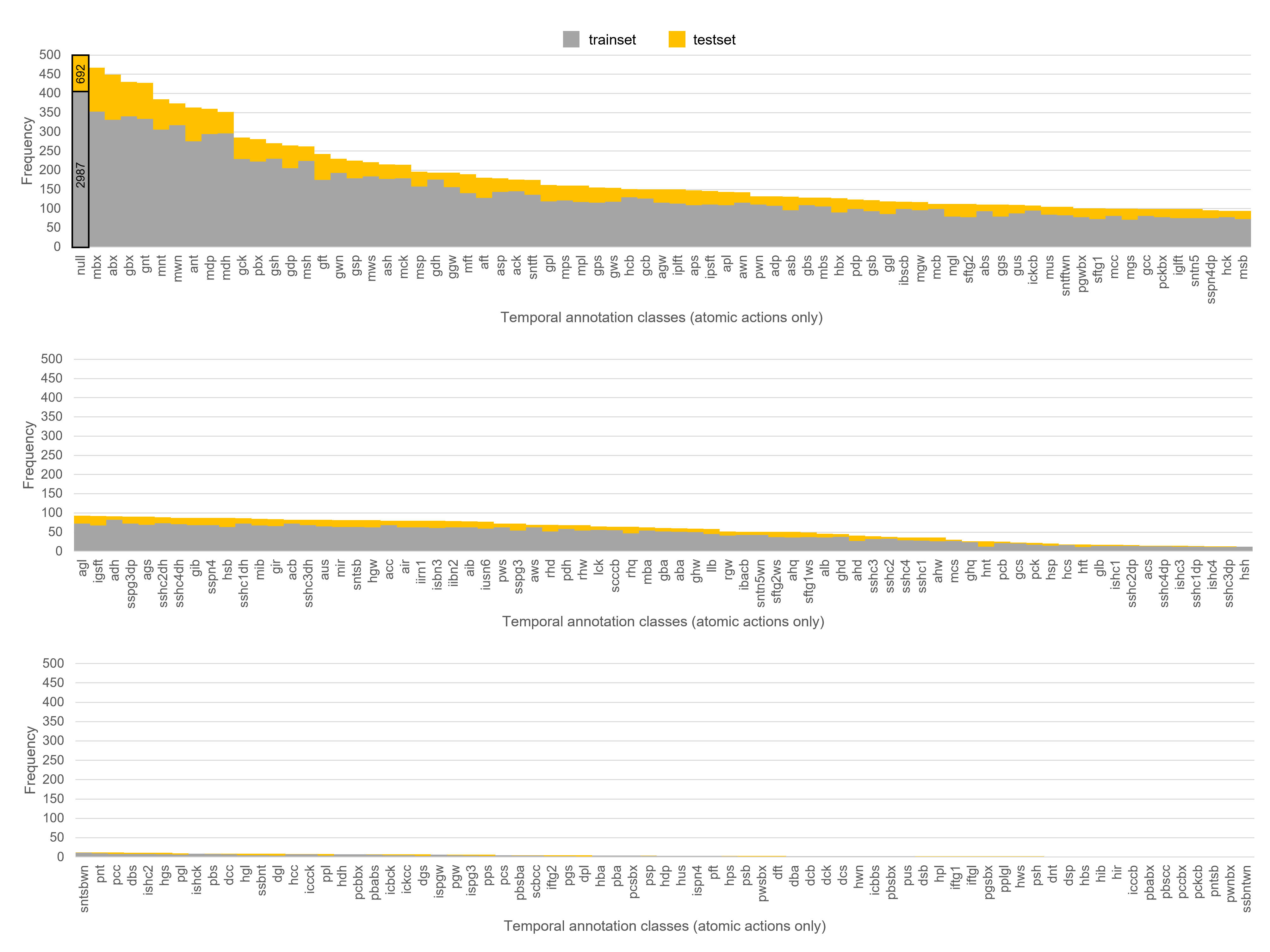}
  \caption{Trainset and testset distribution of the 219 atomic action classes. To show all classes, the diagram is split into three rows. Additionally, to show the distribution better, the frequency axis bound has been reduced, which cuts off the column for the \emph{null} class. Instead, we have manually overwritten the \emph{null} class column with the trainset and testset frequency.}
  \label{figsup12}
\end{figure}

Overall, the dataset contains temporal annotations of 81 primitive task classes and 219 atomic action classes. The trainset and testset were split by subjects to balance data diversity. Figure \ref{figsup11} and Figure \ref{figsup12} show the class distributions of primitive task and atomic action annotations in the trainset and testset, respectively.

\begin{figure}[h!]
  \centering
  \includegraphics[width=\linewidth]{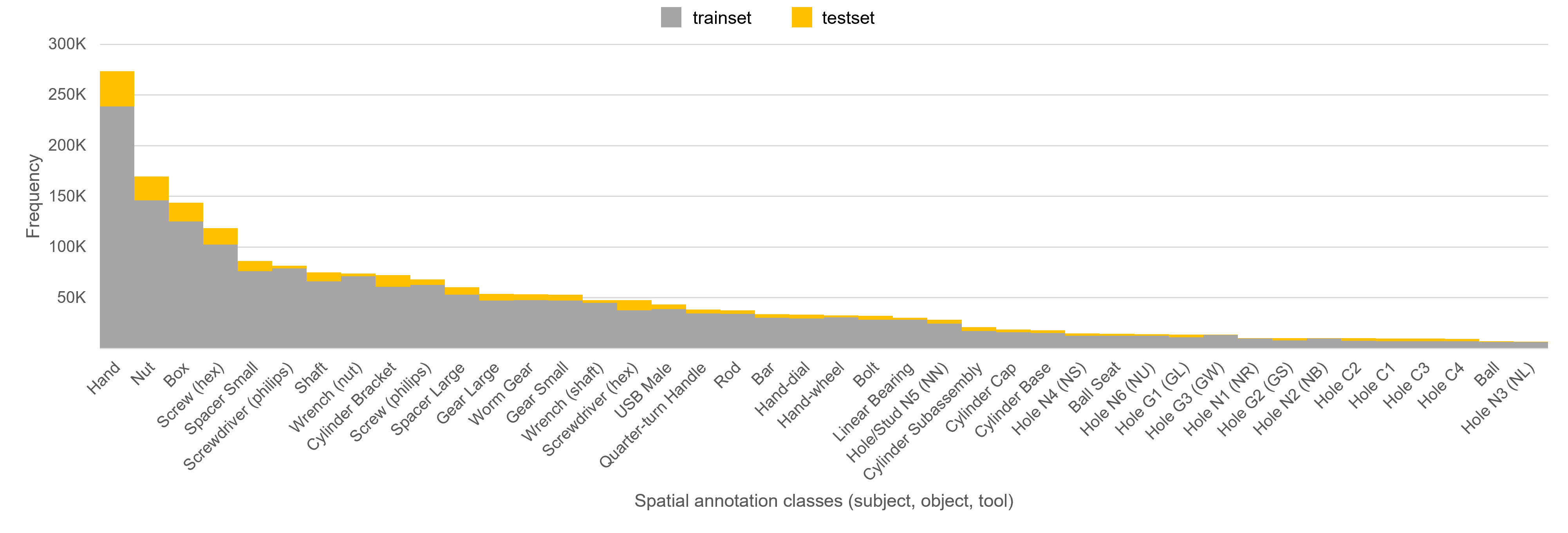}
  \caption{Trainset and testset distribution of the 42 spatial annotation classes. This includes subject, object, and tool.}
  \label{figsup13}
\end{figure}

Overall, the dataset contains spatial annotations of 42 classes. The trainset and testset were split by subjects to balance data diversity. Figure \ref{figsup13} shows the class distributions of spatial annotation classes in the trainset and testset.

\section{Experiment}
\label{sup2}
In this section, we provide the implementation details of the baselines, the results unreleased in the main paper, further discussions on the results, and the licenses of the benchmarked algorithms.

\subsection{Action Recognition}
\begin{figure}[h!]
  \centering
  \includegraphics[width=\linewidth]{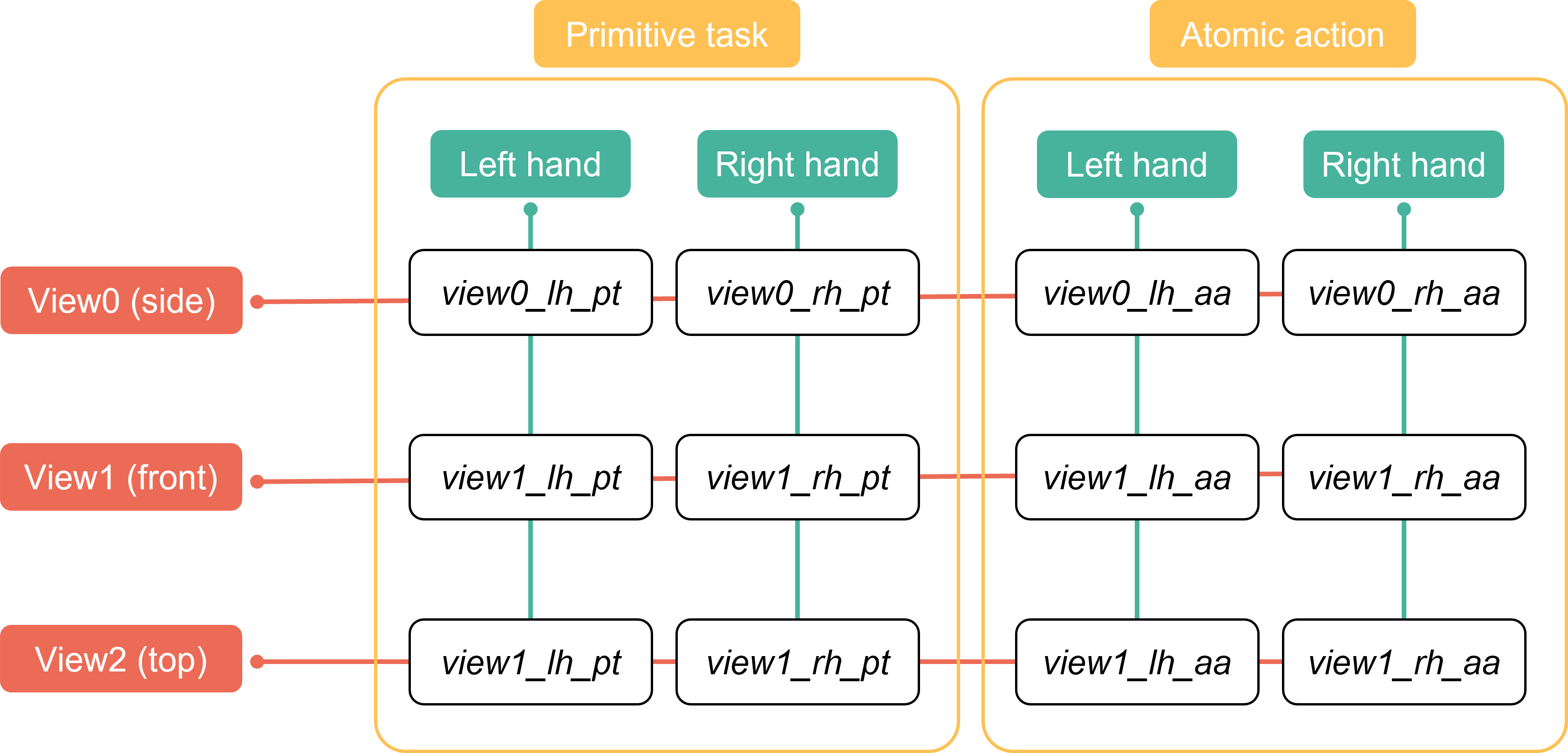}
  \caption{We split the dataset into 12 sub-datasets with three views (side, front, and top), two hands (left and right hands), and two annotation levels (primitive task and atomic action).}
  \label{figsup14}
\end{figure}

We use the MMSkeleton\footnote{https://github.com/open-mmlab/mmskeleton} toolbox to benchmark ST-GCN \cite{Yan2018}; the MMAction2\footnote{https://github.com/open-mmlab/mmaction2} toolbox to benchmark I3D \cite{Carreira2017}, TimeSformer \cite{Bertasius2021}, and MVITv2 \cite{Li2022}; and the original codes to benchmark TSM \cite{Lin2019}. For ST-GCN, we first extracted the upper 26 skeleton joints from each frame as the input. Action clips which consisted of frames where the skeleton could not be extracted, were excluded from reporting the performance. For I3D (rgb), TSM, MVITv2, and TimeSformer, the RGB frames of each clip were used as input. For I3D (flow), we extracted TV-L1 optical flow frames from each clip as input. To compare model performance on different views (side, front, and top), hands (left and right hands) and annotation levels (primitive task and atomic action), we conducted a combinational benchmark, which means we benchmark each model on 12 sub-datasets (see Figure \ref{figsup14}). We report the Top-1 and Top-5 accuracy on these sub-datasets in Table \ref{table2}.

\textbf{ST-GCN}: Following the default parameters from MMSkeleton, we use the SGD optimizer with a dropout of 0.5. The learning rate was initialized as 0.1 and decayed by a factor of 10 after epochs 10 and 50. We sampled all frames as the input. The ST-GCN was pretrained on NTU \cite{Shahroudy2016}, and we finetuned it on our 12 sub-datasets. As the slowest convergence of the 12 sub-datasets was observed around 70 epochs, we set the total training epochs to be 80 with a batch size of 16.

\textbf{TSM}: Following the original paper’s suggestions, we use the SGD optimizer with a dropout of 0.5. The learning rate was initialized as 0.0025 and decayed by a factor of 10 after epochs 20 and 40. 8 frames were uniformly sampled from each clip. The TSM was pretrained on ImageNet \cite{Deng2009}, and we finetuned it on our 12 sub-datasets. As the slowest convergence of the 12 sub-datasets was observed around 40 epochs, we set the total training epochs to be 50 with a batch size of 16. 

\textbf{TimeSformer}: Following the default parameters from MMAction2, we use the SGD optimizer. The learning rate was initialized as 0.005 and decayed by a factor of 10 after epochs 5 and 10. 8 frames were uniformly sampled from each clip. The TimeSformer was pretrained on ImageNet-21K \cite{Deng2009}, and we finetuned it on our 12 sub-datasets. As the slowest convergence of the 12 sub-datasets was observed around 90 epochs, we set the total training epochs to be 100 with a batch size of 8.

\textbf{I3D (rgb) and (flow)}: Following the default parameters from MMAction2, we use the SGD optimizer with a dropout of 0.5. The learning rate was initialized as 0.01 and decayed by a factor of 10 after epochs 40 and 80. 32 frames were uniformly sampled from each clip. I3D takes ResNet50 pretrained on ImageNet-1K \cite{Deng2009} as the backbone, and we finetuned it on our 12 sub-datasets. As the slowest convergence of the 12 sub-datasets was observed around 90 epochs, we set the total training epochs to be 100 with a batch size of 4. 

\textbf{MVITv2}: Following the default parameters from MMAction2, we use the AdamW optimizer with a cosine annealing learning rate with the minimum learning rate of 0.00015. 16 frames were uniformly sampled from each clip. The MVITv2 was pre-trained on Kinetics-400 \cite{Kay2017} via MaskFeat \cite{Wei2022}, and we finetuned it on our 12 sub-datasets. As the slowest convergence of the 12 sub-datasets was observed around 90 epochs, we set the total training epochs to be 100 with a batch size of 4. 

The benchmarking results of action recognition are shown in Table \ref{tablesup2}. We use a single RTX 3090 GPU to train each model, and Table \ref{tablesup3} shows the average training time of each model for each sub-dataset.

\begin{table}[h!]
\centering
\caption{Baselines of action recognition.}
\label{tablesup2}
\resizebox{0.85\textwidth}{!}{%
\begin{tabular}{lllllllllllll}
\hline
\multirow{3}{*}{Method}                     & \multirow{3}{*}{View} & \multicolumn{5}{l}{Primitive Task}                                &  & \multicolumn{5}{l}{Atomic Action}                                 \\ \cline{3-7} \cline{9-13} 
                                            &                       & \multicolumn{2}{l}{Left-Hand} &  & \multicolumn{2}{l}{Right-Hand} &  & \multicolumn{2}{l}{Left-Hand} &  & \multicolumn{2}{l}{Right-Hand} \\ \cline{3-4} \cline{6-7} \cline{9-10} \cline{12-13} 
                                            &                       & Top-1         & Top-5         &  & Top-1          & Top-5         &  & Top-1          & Top-5        &  & Top-1          & Top-5         \\ \cline{1-4} \cline{6-7} \cline{9-10} \cline{12-13} 
\multirow{3}{*}{ST-GCN}                     & Side                  & 40.7          & 61.5          &  & 41.4           & 61.3          &  & 22.2           & 46.0         &  & 21.5           & 44.4          \\
                                            & Front                 & 41.9          & 65.7          &  & 39.3           & 57.7          &  & 21.9           & 46.6         &  & 19.9           & 40.5          \\
                                            & Top                   & 35.8          & 53.4          &  & 35.4           & 46.7          &  & 16.8           & 40.7         &  & 17.8           & 36.9          \\
\multirow{3}{*}{TSM}                        & Side                  & 57.5          & 88.2          &  & 56.8           & 89.7          &  & 38.4           & 67.8         &  & 37.0           & 67.5          \\
                                            & Front                 & 61.5          & 89.3          &  & 57.1           & 85.1          &  & 38.9           & 69.8         &  & 34.3           & 64.6          \\
                                            & Top                   & 64.2          & 88.1          &  & 62.0           & 88.9          &  & 41.6           & 70.8         &  & 39.8           & 69.7          \\
\multirow{3}{*}{TimeSformer}                & Side                  & 53.8          & 85.8          &  & 50.6           & 85.7          &  & 36.8           & 69.7         &  & 31.8           & 64.7          \\
                                            & Front                 & 50.8          & 84.4          &  & 48.9           & 80.5          &  & 36.8           & 68.0         &  & 32.8           & 62.9          \\
                                            & Top                   & 51.7          & 86.0          &  & 55.9           & 87.0          &  & 39.1           & 68.7         &  & 39.3           & 70.8          \\
\multirow{3}{*}{I3D (flow)}                 & Side                  & 38.6          & 50.6          &  & 37.0           & 44.9          &  & 23.8           & 46.8         &  & 23.8           & 45.3          \\
                                            & Front                 & 39.1          & 54.7          &  & 37.0           & 45.1          &  & 23.7           & 48.1         &  & 23.5           & 46.5          \\
                                            & Top                   & 39.4          & 57.9          &  & 37.3           & 48.7          &  & 22.6           & 45.3         &  & 23.9           & 45.9          \\
\multirow{3}{*}{I3D (rgb)}                  & Side                  & 54.9          & 82.5          &  & 51.8           & 83.7          &  & 38.2           & 72.0         &  & 34.0           & 66.8          \\
                                            & Front                 & 52.8          & 83.6          &  & 51.6           & 82.9          &  & 41.6           & 73.5         &  & 35.6           & 66.0          \\
                                            & Top                   & 54.4          & 85.0          &  & 57.6           & 84.0          &  & 41.3           & 70.3         &  & 41.2           & 71.3          \\
\multirow{3}{*}{I3D (both)}                 & Side                  & 32.2          & 45.7          &  & 51.1           & 85.2          &  & 40.8           & 75.6         &  & 37.6           & 71.4          \\
                                            & Front                 & 53.2          & 83.6          &  & 49.7           & 84.4          &  & 44.0           & 75.9         &  & 39.6           & 71.3          \\
                                            & Top                   & 57.7          & 85.0          &  & 57.8           & 85.6          &  & 44.1           & 73.5         &  & 44.4           & 75.9          \\
\multicolumn{1}{c}{\multirow{3}{*}{MVITv2}} & Side                  & 58.5          & 85.2          &  & 57.8           & 85.2          &  & 48.5           & 76.5         &  & 41.8           & 70.8          \\
\multicolumn{1}{c}{}                        & Front                 & 63.1          & 86.6          &  & 55.9           & 81.6          &  & 48.3           & 76.4         &  & 41.9           & 70.1          \\
\multicolumn{1}{c}{}                        & Top                   & 62.9          & 87.1          &  & 62.5           & 85.4          &  & 48.3           & 76.5         &  & 44.9           & 72.8          \\ \hline
\end{tabular}%
}
\end{table}

\begin{table}[]
\centering
\caption{Training efficiency of ST-GCN, TSM, TimeSformer, I3D, and MVITv2.}
\label{tablesup3}
\resizebox{\textwidth}{!}{%
\begin{tabular}{ccclcccccc}
\hline
\multicolumn{3}{c}{Dataset}                                           &  & \multicolumn{6}{c}{Average training time per epoch (min)}     \\ \cline{1-3} \cline{5-10} 
View                   & Hand                        & Task level     &  & ST-GCN & TSM  & TimeSformer & I3D (flow) & I3D (rgb) & MVITv2 \\ \cline{1-3} \cline{5-10} 
\multirow{4}{*}{Side}  & \multirow{2}{*}{Left hand}  & Primitive task &  & 1.65   & 1.3  & 6.12        & 3.3        & 5.83      & 11.12  \\
                       &                             & Atomic action  &  & 5.55   & 2.6  & 14.42       & 10.82      & 10.02     & 24.9   \\
                       & \multirow{2}{*}{Right hand} & Primitive task &  & 1.73   & 1.4  & 4.2         & 4.22       & 5.72      & 6.95   \\
                       &                             & Atomic action  &  & 5.38   & 4.48 & 12.85       & 9.12       & 11.73     & 23.55  \\
\multirow{4}{*}{Front} & \multirow{2}{*}{Left hand}  & Primitive task &  & 1.73   & 1.33 & 3.93        & 4.15       & 5.88      & 11.15  \\
                       &                             & Atomic action  &  & 5.72   & 4.5  & 21.4        & 9.63       & 12.23     & 25.37  \\
                       & \multirow{2}{*}{Right hand} & Primitive task &  & 1.82   & 1.22 & 4.22        & 2.48       & 4.68      & 6.98   \\
                       &                             & Atomic action  &  & 5.65   & 4.27 & 12.82       & 7.02       & 11.18     & 26.58  \\
\multirow{4}{*}{Top}   & \multirow{2}{*}{Left hand}  & Primitive task &  & 0.71   & 1.38 & 4.08        & 5.25       & 5.55      & 11.5   \\
                       &                             & Atomic action  &  & 3.01   & 4.75 & 14.3        & 10.05      & 11.57     & 24.05  \\
                       & \multirow{2}{*}{Right hand} & Primitive task &  & 0.65   & 1.4  & 4.17        & 4.47       & 2.8       & 8.33   \\
                       &                             & Atomic action  &  & 2.43   & 4.57 & 12.8        & 7.07       & 10.93     & 24.03  \\ \hline
\end{tabular}%
}
\end{table}

\subsection{Action Segmentation}

We benchmark three action segmentation algorithms: MS-TCN, DTGRM, and BCN, and report the frame-wise accuracy (Acc), segmental edit distance (Edit) and segmental F1 score at overlapping thresholds 10\% in Table \ref{table4}. Before benchmarking, we extract I3D features for each frame as the input of the action segmentation algorithms. We use the Pytorch version of the I3D implementation\footnote{https://github.com/piergiaj/pytorch-i3d} and the pretrained model on ImageNet \cite{Deng2009} and Kinetics \cite{Kay2017}. For action segmentation, we also conducted a combinational benchmark.

\textbf{MS-TCN}: We follow the model settings provided by \cite{Farha2019}. More specifically, we use the Adam optimizer with a fixed learning rate of 0.0005, dropout of 0.5 and sampling rate of 1 (taking all frames into the network). As the slowest convergence of the 12 sub-datasets was observed around 800 epochs, we set the total training epochs to be 1000 with a batch size of 10.

\textbf{DTGRM}: We follow the model settings provided by \cite{Wang2021}. More specifically, we use the Adam optimizer with a fixed learning rate of 0.0005, dropout of 0.5 and sampling rate of 1. As the slowest convergence of the 12 sub-datasets was observed around 800 epochs, we set the total training epochs to be 1000 with a batch size of 16.

\textbf{BCN}: We follow the model settings provided by \cite{Wang2020}. More specifically, we use the Adam optimizer with the learning rate of 0.001 for the first 30 epochs and 0.0001 for the rest epochs, dropout of 0.5 and sampling rate of 1. As the slowest convergence of the 12 sub-datasets was observed around 200 epochs, we set the total training epochs to be 300 with a batch size of 1.

The benchmarking results of action segmentation are shown in Table \ref{tablesup4}. We use a single RTX 3090 GPU to train each model, and Table \ref{tablesup5} shows the average training time of each model for each sub-dataset.

\begin{table}[]
\centering
\caption{Baselines of Action Segmentation.}
\label{tablesup4}
\resizebox{\textwidth}{!}{%
\begin{tabular}{ccccclccclccclccc}
\hline
\multirow{3}{*}{Method} & \multirow{3}{*}{View} & \multicolumn{7}{c}{Primitive task}                                &  & \multicolumn{7}{c}{Atomic Action}                                 \\ \cline{3-9} \cline{11-17} 
                        &                       & \multicolumn{3}{c}{Left hand} &  & \multicolumn{3}{c}{Right hand} &  & \multicolumn{3}{c}{Left hand} &  & \multicolumn{3}{c}{Right hand} \\ \cline{3-5} \cline{7-9} \cline{11-13} \cline{15-17} 
                        &                       & F1       & Edit     & Acc     &  & F1       & Edit     & Acc      &  & F1       & Edit     & Acc     &  & F1       & Edit     & Acc      \\ \cline{1-5} \cline{7-9} \cline{11-13} \cline{15-17} 
\multirow{3}{*}{MS-TCN} & Side                  & 37.6     & 37.4     & 41.2    &  & 31.1     & 32.5     & 37.4     &  & 35.3     & 32.1     & 40.9    &  & 29.2     & 31.0     & 32.6     \\
                        & Front                 & 35.2     & 36.3     & 38.8    &  & 36.7     & 36.2     & 39.3     &  & 34.1     & 31.2     & 41.1    &  & 29.3     & 31.1     & 33.4     \\
                        & Top                   & 37.1     & 38.9     & 40.4    &  & 36.1     & 35.6     & 41.3     &  & 35.9     & 34.1     & 40.8    &  & 35.1     & 34.6     & 37.8     \\
\multirow{3}{*}{DTGRM}  & Side                  & 38.5     & 36.5     & 40.9    &  & 35.9     & 35.2     & 37.6     &  & 33.7     & 30.7     & 39.3    &  & 27.8     & 28.2     & 30.3     \\
                        & Front                 & 38.5     & 37.2     & 39.0    &  & 38.8     & 39.6     & 40.5     &  & 34.0     & 33.6     & 39.7    &  & 27.6     & 27.8     & 31.5     \\
                        & Top                   & 40.4     & 38.8     & 40.8    &  & 38.7     & 37.0     & 41.2     &  & 35.1     & 33.6     & 40.5    &  & 34.0     & 31.9     & 37.6     \\
\multirow{3}{*}{BCN}    & Side                  & 43.1     & 40.4     & 43.7    &  & 38.6     & 36.3     & 42.4     &  & 21.3     & 18.0     & 39.5    &  & 20.5     & 18.9     & 34.1     \\
                        & Front                 & 44.4     & 43.1     & 44.4    &  & 41.3     & 37.0     & 44.0     &  & 17.2     & 14.4     & 39.5    &  & 22.9     & 20.7     & 34.3     \\
                        & Top                   & 43.5     & 40.7     & 44.3    &  & 44.0     & 40.7     & 43.7     &  & 16.8     & 15.3     & 40.1    &  & 23.4     & 20.6     & 35.5     \\ \hline
\end{tabular}%
}
\end{table}

\begin{table}[]
\centering
\caption{Training efficiency of MS-TCN, DTGRM and BCN.}
\label{tablesup5}
\resizebox{0.7\textwidth}{!}{%
\begin{tabular}{ccclccc}
\hline
\multicolumn{3}{c}{Dataset}                                           &  & \multicolumn{3}{c}{Average training time per epoch (sec)} \\ \cline{1-3} \cline{5-7} 
View                   & Hand                        & Task level     &  & MS-TCN             & DTGRM             & BCN              \\ \cline{1-3} \cline{5-7} 
\multirow{4}{*}{Side}  & \multirow{2}{*}{Left hand}  & Primitive task &  & 8.24               & 18.66             & 16.35            \\
                       &                             & Atomic action  &  & 8.37               & 19.42             & 16.50            \\
                       & \multirow{2}{*}{Right hand} & Primitive task &  & 8.86               & 20.01             & 16.26            \\
                       &                             & Atomic action  &  & 8.66               & 20.41             & 16.51            \\
\multirow{4}{*}{Front} & \multirow{2}{*}{Left hand}  & Primitive task &  & 8.04               & 19.44             & 16.31            \\
                       &                             & Atomic action  &  & 8.01               & 19.82             & 16.38            \\
                       & \multirow{2}{*}{Right hand} & Primitive task &  & 8.31               & 20.05             & 16.24            \\
                       &                             & Atomic action  &  & 8.45               & 19.12             & 16.56            \\
\multirow{4}{*}{Top}   & \multirow{2}{*}{Left hand}  & Primitive task &  & 7.81               & 19.44             & 16.39            \\
                       &                             & Atomic action  &  & 7.97               & 19.44             & 16.42            \\
                       & \multirow{2}{*}{Right hand} & Primitive task &  & 8.23               & 18.70             & 16.31            \\
                       &                             & Atomic action  &  & 8.30               & 19.27             & 16.51            \\ \hline
\end{tabular}%
}
\end{table}

\subsection{Object Detection}
We benchmark three object detection algorithms: Faster-RCNN \cite{Ren2017}, YOLOv5 \cite{Jain} and DINO \cite{Zhang2022a} with different backbone networks. The results have been reported in the main paper. Therefore, we only discuss the implementation details here. We train Faster-RCNN and DINO using the implementation provided by the MMDetection \cite{Chen2019} and train YOLOv5 using the implementation provided by the MMYOLO\footnote{https://github.com/open-mmlab/mmyolo}.

\textbf{Faster-RCNN}: We train Faster-RCNN with three backbone networks: ResNet50, ResNet101, and ResNext101. All the networks have been pretrained on the coco\_2017\_train dataset \cite{Lin2014} and finetuned on our dataset. Following the default setting provided by MMDetection, we use the SGD optimizer with a momentum of 0.9 and weight decay of 0.0001. The learning rate was initialized as 0.02 and decayed by a factor of 10 at epochs 8 and 11. As the slowest convergence of the three models was observed around 14 epochs, we set the total training epochs to be 20. We set the batch size as 4, 1, and 5, respectively, for ResNet50, ResNet101, and ResNext101.

\textbf{YOLOv5}: We train YOLOv5-small and YOLOv5-large using MMDetection. These two models have been pretrained on the coco\_2017\_train dataset, and finetuned on our dataset. Following the default setting provided by MMDetection, we use the SGD optimizer with a momentum of 0.937, weight decay of 0.0005 for both models. The linear learning rate with base learning rate of 0.0025 and factor of 0.01 was applied to YOLOv5-small. The linear learning rate with base learning rate of 0.0025 and factor of 0.1 was applied to YOLOv5-large. We set the total training epochs to be 100 epochs with a batch size of 32 and 50 epochs with a batch size of 10, respectively, for YOLOv5-small and YOLOv5-large to ensure convergence.

\textbf{DINO}: We benchmark the DINO model with the Swin-large network as the backbone. The model has been pretrained on the coco\_2017\_train dataset, and finetuned on our dataset. Following the default setting provided by MMDetection, we use the AdamW optimizer with a learning rate of 0.0001 and weight decay of 0.0001. As the convergence was observed around 6 epochs, we set the total training epochs to be 10 with a batch size of 1.

We use single RTX 3090 GPU to train each model, and Table \ref{tablesup6} shows the average training time of each model.

\begin{table}[!h]
\centering
\caption{Training efficiency of Faster-RCNN, YOLOv5 and DINO.}
\label{tablesup6}
\resizebox{0.6\textwidth}{!}{%
\begin{tabular}{ccc}
\hline
\multicolumn{2}{c}{Method}                & Average training time per epoch (min) \\ \hline
\multirow{3}{*}{Faster-RCNN} & ResNet50   & 446.9                                 \\
                             & ResNet101  & 197.0                                 \\
                             & ResNext101 & 668.8                                 \\ \cline{1-2}
YOLOv5-s                     & DarkNet    & 39.5                                  \\
YOLOv5-l                     & DarkNet    & 94.2                                  \\
DINO                         & Swin-L     & 1592.3                                \\ \hline
\end{tabular}%
}
\end{table}

\subsection{Multi-Object Tracking}
In this paper, we focus on tracking-by-detection methods because, normally, tracking-by-detection methods perform better than joint-detection-association methods \cite{Luo2021}. Since we already benchmarked the object detection methods, we only need to test the SOTA trackers. We benchmark SORT \cite{Bewley2016} and ByteTrack \cite{Zhang2022} trackers on the detection results of DINO and ground truth annotations, respectively. The results have been reported in the main paper. Since the trackers are not neural networks, we do not need to train them and explain the implementation details. We always use the default parameters of the algorithm. For more details, please refer to the papers \cite{Bewley2016,Zhang2022} and their GitHub repositories.

\subsection{Discussion}

In this section, we further discuss the results from the above experiments and analyze a prevalent problem of video understanding – occlusion.

\subsubsection{General Discussion}
\label{sup2.5.1}
\textbf{Action recognition}: We found the Top-1 accuracy of primitive task recognition is 15.6\% higher on average than atomic action recognition, and the atomic action recognition performance of the left hand is 2.4\% higher on average than the right hand.  One possible reason behind these two observations can be occlusion since (1) primitive task recognition is less influenced by occlusion because it can rely on the key motion or relevant object recognition; and (2) the left hand is less occluded because the side-view camera is mounted on the left-side of the participant.

\textbf{Action segmentation}: We found (1) the frame-wise accuracy (Acc) of atomic action segmentation is 4\% lower on average than primitive task segmentation, as atomic actions have higher diversity and current methods face under-segmentation issues (refer to the main paper); and (2) on the atomic action level, the Acc of the left hand is 6\% higher on average than the right hand, where one possible reason could be that the left hand is less occluded.

\begin{figure}[h!]
  \centering
  \includegraphics[width=\linewidth]{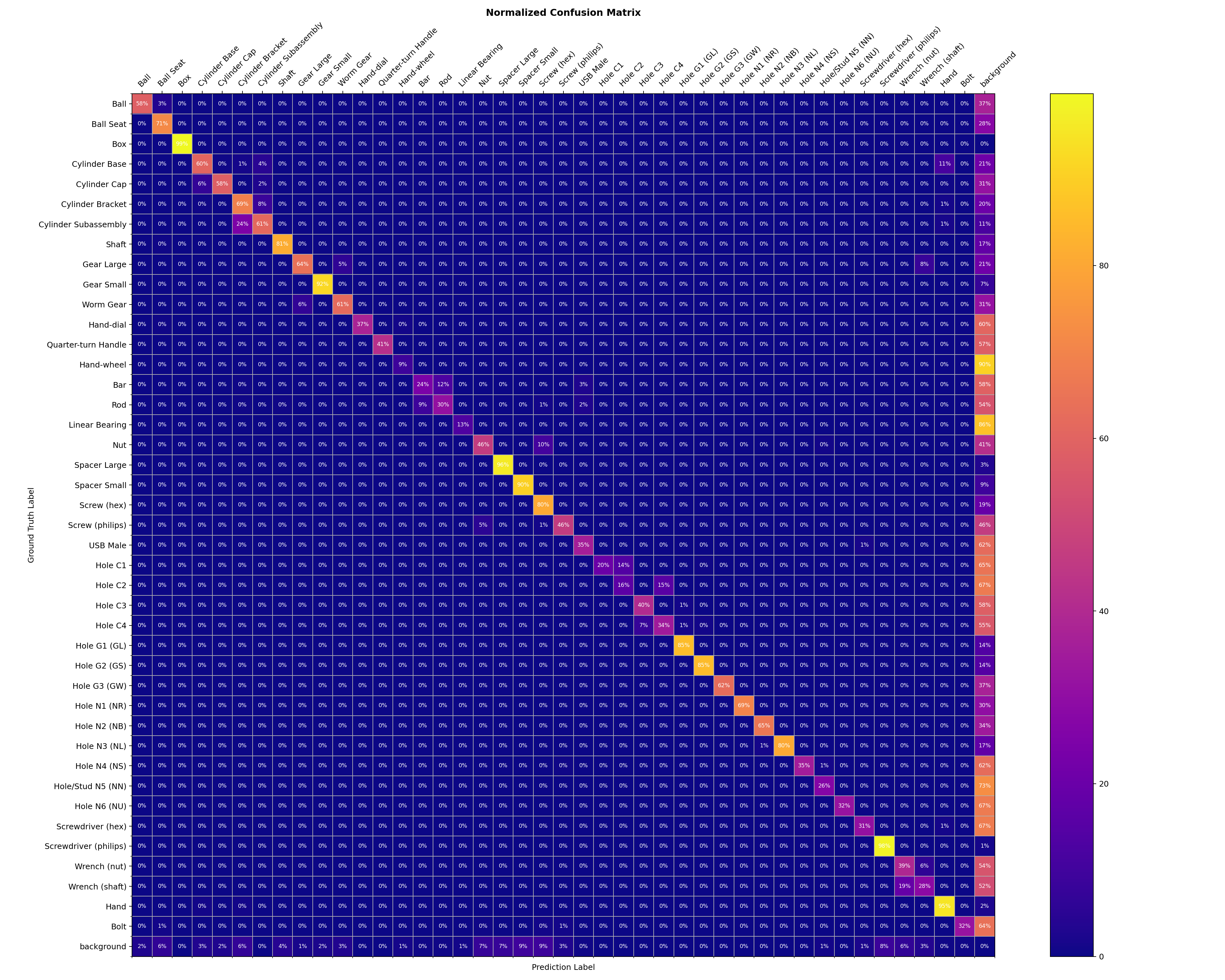}
  \caption{Confusion matrix of object detection results from DINO.}
  \label{figsup15}
\end{figure}

\textbf{Object detection}: From Table 4 of the main paper, we found that (1) the large-scale end-to-end Transformer based model (DINO) performs the best, and the traditional two-stage method (Faster-RCNN) has better performance on small objects but worse performance on large objects than the one-stage method (YOLOv5), which is consistent with the conclusion of \cite{Zhao2019}; (2) current methods still face great challenges in small object detection, as the best model only has 27.4\% average precision on small object detection; and (3) recognizing objects with same/similar appearances but different sizes is challenging (see Figure \ref{figsup15}, e.g., Bar and Rod, Hole C1-C4, and two Wrenches).

\begin{figure}[h!]
  \centering
  \includegraphics[width=\linewidth]{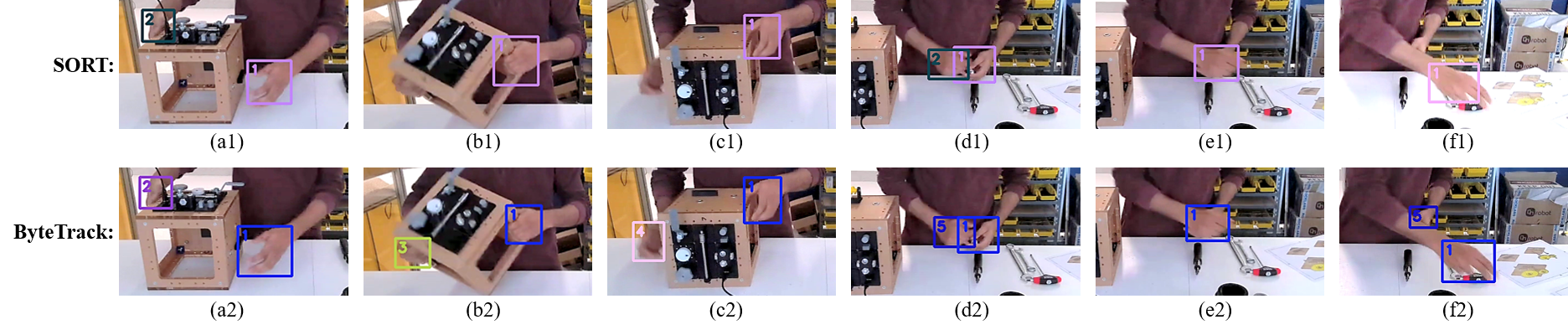}
  \caption{Confusion matrix of object detection results from DINO.}
  \label{figsup16}
\end{figure}

\textbf{Multi-object detection}: From Table 5 of the main paper, we found that (1) object detection performance is the decisive factor in tracking performance; (2) with perfect detection results, even the simple tracker (SORT) can achieve good tracking results, as SORT has 94.5\% multi-object tracking accuracy on the ground truth object bounding boxes; and (3) ByteTrack can track blurred and occluded objects better (comparing b1-2, c1-2, and f1-2 in Figure \ref{figsup16}) due to taking low-confidence detection results into association, but it generates more ID switches (IDS) (seeing a2-f2 in Figure \ref{figsup16}) due to the preference of creating new tracklets.

\subsubsection{Occlusion Analysis}
From the discussion in Section \ref{sup.5.1}, we can see occlusion is a prevalent problem of video understanding. Therefore, we further explore the impact of occlusion on video understanding tasks in this Section. Table \ref{tablesup7} reports the average results over two hands of action recognition and segmentation on three views and the combined view (Com). We fuse the features from three views before the softmax layer to evaluate the performance of the combined view. The results show the significant benefits of combining three views which offers a viable solution for mitigating occlusion challenges in industrial settings.

\begin{table}[h!]
\centering
\caption{Performance of action recognition and segmentation on three views and the combined view.}
\label{tablesup7}
\resizebox{0.9\textwidth}{!}{%
\begin{tabular}{cccccccccccccc}
\hline
\multirow{3}{*}{View} & \multicolumn{7}{c}{Action Segmentation (BCN)}                                                             &           & \multicolumn{5}{c}{Action Recognition (MVITv2)}                                    \\ \cline{2-8} \cline{10-14} 
                      & \multicolumn{3}{c}{Primitive task}            &           & \multicolumn{3}{c}{Atomic action}             &           & \multicolumn{2}{c}{Primitive task} &           & \multicolumn{2}{c}{Atomic action} \\ \cline{2-4} \cline{6-8} \cline{10-11} \cline{13-14} 
                      & F1            & Edit          & Acc           &           & F1            & Edit          & Acc           &           & Top-1             & Top-5          &           & Top-1           & Top-5           \\ \cline{1-4} \cline{6-8} \cline{10-11} \cline{13-14} 
Side                  & 40.9          & 38.4          & 43.1          &           & 20.9          & 18.5          & 36.8          &           & 58.2              & 85.2           &           & 45.2            & 73.7            \\
Front                 & 42.9          & 40.1          & 44.2          &           & 20.1          & 17.6          & 36.9          &           & 59.5              & 84.1           &           & 45.1            & 73.3            \\
Top                   & 43.8          & 40.7          & 44            &           & 20.1          & 18.0          & 37.8          &           & 62.7              & 86.3           &           & 46.6            & 74.7            \\
Com                   & \textbf{44.6} & \textbf{45.9} & \textbf{47.2} & \textbf{} & \textbf{41.7} & \textbf{35.9} & \textbf{44.5} & \textbf{} & \textbf{64.0}     & \textbf{89}    & \textbf{} & \textbf{50.8}   & \textbf{80.9}   \\ \hline
\end{tabular}%
}
\end{table}

Figure \ref{figsup16} shows the impact of occlusion on tracking and reidentification via visualizing SORT and ByteTrack tracking results on sampled ground truth object annotations. To quantitatively analyze the occlusion problem, we design two metrics: occlusion duration (OD) and occlusion frequency (OF). Given a video of \(n\) frames \(v=[f_1,\ldots,f_n]\), the observation of object \(k\) is denoted as \(O_k=[o_t^k,o_{t+1}^k,\ldots,o_{t+m}^k]\), where \(t\) and \(t+m\) are the frame numbers that object \(k\) first, and last appear, respectively. \(o_j^k=\{0,1\}\), where 0 denotes observed, and 1 denotes unobserved. \(\text{OD}_k=\frac{1}{m}\sum_{j=t}^{j=t+m}o_j^k\) and \(\text{OF}_k=\frac{1}{2}\sum_{j=t}^{j=t+m-1}|o_{j+1}^k-o_j^k|\). \(\text{OD}_k\) and \(\text{OF}_k\) describe the occluded duration and occluded frequency of object \(k\) in a video. We calculate the average OD and OF over every object in our testing dataset and compare the results with the tracking results on ground truth object annotations in Table \ref{table8}. Table \ref{tablesup8} shows a negative correlation between mOD and mOF with MOTA and IDS, which is also consistent with the findings in Figure \ref{figsup16}. We envision OD and OF will serve as effective occlusion evaluation tools for developing better object association modules and reidentification modules in MOT.

\begin{table}[]
\centering
\caption{Comparison between tracking results and occlusion metrics on three views.}
\label{tablesup8}
\resizebox{0.6\textwidth}{!}{%
\begin{tabular}{ccccccc}
\hline
View                   & Method                        & MOTA                       & IDF1                       & IDS                       & mOD                     & mOF                  \\ \hline
\multirow{2}{*}{Side}  & SORT                          & 93.5\%                     & 66.5\%                     & 58.3                      & \multirow{2}{*}{18.7\%} & \multirow{2}{*}{4.1} \\
                       & ByteTrack                     & 98.5\%                     & 68.4\%                     & 124.5                     &                         &                      \\ \hline
\multirow{2}{*}{Front} & SORT                          & 95.3\%                     & 72.1\%                     & 48.2                      & \multirow{2}{*}{12.1\%} & \multirow{2}{*}{2.9} \\
                       & ByteTrack                     & 98.7\%                     & 67.8\%                     & 118.7                     &                         &                      \\ \hline
\multirow{2}{*}{Top}   & SORT                          & 94.7\%                     & 68.6\%                     & 57.8                      & \multirow{2}{*}{14.7\%} & \multirow{2}{*}{5.3} \\
                       & \multicolumn{1}{l}{ByteTrack} & \multicolumn{1}{l}{98.4\%} & \multicolumn{1}{l}{66.3\%} & \multicolumn{1}{l}{121.5} &                         &                      \\ \hline
\end{tabular}%
}
\end{table}

\subsection{Licenses of the benchmarked algorithms}
The licenses of the benchmarked algorithms are listed in Table \ref{tablesup9}.

\begin{table}[h!]
\centering
\caption{Licenses of the benchmarked algorithms.}
\label{tablesup9}
\resizebox{0.55\textwidth}{!}{%
\begin{tabular}{cc}
\hline
Algorithm   & License                                     \\ \hline
MMSkeleton  & Apache License 2.0                          \\
ST-GCN      & BSD 2-Clause "Simplified" License           \\
MMAction2   & Apache License 2.0                          \\
TSM         & MIT                                         \\
TimeSFormer & Attribution-NonCommercial 4.0 International \\
I3D         & Apache License 2.0                          \\
MVITv2      & Apache License 2.0                          \\
MS-TCN      & MIT                                         \\
DTGRM       & MIT                                         \\
BCN         & MIT                                         \\
MMDetection & Apache License 2.0                          \\
Faster-RCNN & MIT                                         \\
DINO        & Apache License 2.0                          \\
MMYOLO      & GNU General Public License v3.0             \\
YOLOv5      & GNU Affero General Public License v3.0      \\
SORT        & GNU General Public License v3.0             \\
ByteTrack   & MIT                                         \\ \hline
\end{tabular}%
}
\end{table}

\section{Dataset Bias and Societal Impact}
\label{sup3}
Our objective is to construct a dataset that can represent interesting and challenging problems in real-world industrial assembly scenarios. Based on this objective, we developed the Generic Assembly Box that encompasses standard and non-standard parts widely used in industry and requires typical industrial tools to assemble. However, there is still a gap between our dataset and the real-world industrial assembly scenarios. The challenges lie in: 
\begin{enumerate}
  \item[1)] the existence of numerous unique assembly actions, countless parts, and tools in the industry;
  \item[2)] the vast diversity of operating environments in the industry;
  \item[3)] various agents and multi-agent collaborative assembly scenarios in the industry.
\end{enumerate}

Therefore, additional efforts would be needed to apply the models trained on our dataset to real-world industrial applications. We hope the fine-grained annotations of this dataset can advance the technological breakthrough in comprehensive assembly knowledge understanding from videos. Then, the learned knowledge can benefit various real-world applications, such as robot skill learning, human-robot collaboration, assembly process monitoring, assembly task planning, and quality assurance. We hope this dataset can contribute to technological advancements facilitating the development of smart manufacturing, enhancing production efficiency, and reducing the workload and stress on workers.

\section{Ethics Approval}
\label{sup4}
HA-ViD was collected with ethics approval from the University of Auckland Human Participants Ethics Committee. The Reference Number is 21602. All participants were sent a Participant Information Sheet and Consent Form\footnote{The participant consent form is available at: \url{https://www.dropbox.com/sh/ekjle5bwoylmdcf/AACLd_NqT3p2kxW7zLvvauPta?dl=0}} prior to the collection session. We confirmed that they had agreed to and signed the Consent form before proceeding with any data collection. 

\section{Data Documentation}
\label{sup5}
We follow the datasheet proposed in \cite{Gebru2018} for documenting our HA-ViD dataset: 

1. Motivation

(a) For what purpose was the dataset created? 

This dataset was created to understand comprehensive assembly knowledge from videos. The previous assembly video datasets fail to (1) represent real-world industrial assembly scenarios, (2) capture natural human behaviors (varying efficiency, alternative routes, pauses and errors) during procedural knowledge acquisition, (3) follow a consistent annotation protocol that aligns with human and robot assembly comprehension.

(b) Who created the dataset, and on behalf of which entity? 

This dataset was created by Hao Zheng, Regina Lee and Yuqian Lu. At the time of creation, Hao and Regina were PhD students at the University of Auckland, and Yuqian was a senior lecturer at the University of Auckland.

(c) Who funded the creation of the dataset? 

The creation of this dataset was partially funded by The University of Auckland FRDF New Staff Research Fund (No. 3720540).

(d) Any other Comments? 

None. 

2. Composition 

(a) What do the instances that comprise the dataset represent? 

For the video dataset, each instance is a video clip recording a participant assembling one of the three plates of the designed Generic Assembly Box. Each instance consists of two-level temporal annotations: primitive task and atomic action, and spatial annotations, which means the bounding boxes for subjects, objects, and tools. 

(b) How many instances are there in total? 

We recorded 3222 videos over 86.9 hours, totaling over 1.5M frames. To ensure annotation quality, we manually labeled temporal annotations for 609 plate assembly videos and spatial annotations for over 144K frames. 

(c) Does the dataset contain all possible instances, or is it a sample (not necessarily random) of instances from a larger set? 

Yes, the dataset contains all possible instances.

(d) What data does each instance consist of? 

See 2. (a). 

(e) Is there a label or target associated with each instance?

See 2. (a).

(f) Is any information missing from individual instances? 

No. 

(g) Are relationships between individual instances made explicit? 

Yes, each instance (video clip) contains one participant performing one task (assembling one of the three plates of the designed Generic Assembly Box.)

(h) Are there recommended data splits? 

For action recognition and action segmentations, we provide two data splits: trainset and testset.

For object detection and multi-object tracking, we provide another two data splits: trainset and testset. 

Refer to Section \ref{sup2.4} for details.

(i) Are there any errors, sources of noise, or redundancies in the dataset? 

Given the scale of the dataset and complexity in annotation, it is possible that some ad-hoc errors exist in our annotations. However, we have given our best efforts (via human checks and quality checking code scripts) in examining manually labelled annotations to minimize these errors. 

(j) Is the dataset self-contained, or does it link to or otherwise rely on external resources (e.g., websites, tweets, other datasets)? 

The dataset is self-contained. 

(k) Does the dataset contain data that might be considered confidential (e.g., data that is protected by legal privilege or by doctor-patient confidentiality, data that includes the content of individuals’ non-public communications)? 

No. 

(l) Does the dataset contain data that, if viewed directly, might be offensive, insulting, threatening, or might otherwise cause anxiety? 

No. 

(m) Does the dataset relate to people? 

Yes, all videos are recordings of human assembly activities, and all annotations are related to the activities. 

(n) Does the dataset identify any subpopulations (e.g., by age, gender)? 

No. Our participants have different ages and genders. But our dataset does not identify this information. To ensure this, we have blurred participants’ faces in the released videos.

(o) Is it possible to identify individuals (i.e., one or more natural persons), either directly or indirectly (i.e., in combination with other data) from the dataset? 

No, as explained in 2. (n), we have blurred participants’ faces in the released videos. 

(p) Does the dataset contain data that might be considered sensitive in any way (e.g., data that reveals racial or ethnic origins, sexual orientations, religious beliefs, political opinions or union memberships, or locations; financial or health data; biometric or genetic data; forms of government identification, such as social security numbers; criminal history)? 

No. 

(q) Any other comments? 

None. 

3. Collection Process 

(a) How was the data associated with each instance acquired?

For each video instance, we provide temporal annotations and spatial annotations. We follow HR-SAT to create temporal annotations to ensure the annotation consistency. The temporal annotations were manually created and checked by our researchers. The spatial annotations were manually created by postgraduate students at the University of Auckland, who were trained by one of our researchers to ensure the annotation quality.

(b) What mechanisms or procedures were used to collect the data (e.g., hardware apparatus or sensor, manual human curation, software program, software API)?

Data were collected on three Azure Kinect RGB+D cameras via live video capturing while a participant is performing the assembly actions, and we manually labeled all the annotations.

(c) If the dataset is a sample from a larger set, what was the sampling strategy (e.g., deterministic, probabilistic with specific sampling probabilities)? 

No, we created a new dataset. 

(d) Who was involved in the data collection process (e.g., students, crowdworkers, contractors) and how were they compensated (e.g., how much were crowdworkers paid)? 

For video recordings, volunteer participants were rewarded gift cards worth NZ\$50.00 upon completion of the 2-hour data collection session.

For data annotations, we contracted students at the University of Auckland, and they were paid at a rate of NZ\$23.00 per hour. 

(e) Over what timeframe was the data collected? 

The videos were recorded during August to September of 2022, and the annotations were made during October of 2022 to March of 2023.

(f) Were any ethical review processes conducted (e.g., by an institutional review board)? 

Yes, we obtained ethics approval from the University of Auckland Human Participants Ethics Committee. More information can be found in Section \ref{sup4}.

(g) Does the dataset relate to people? 

Yes, we recorded the process of people assembling the Generic Assembly Box.

(h) Did you collect the data from the individuals in question directly, or obtain it via third parties or other sources (e.g., websites)? 

We collected the data from the individuals in question directly.

(i) Were the individuals in question notified about the data collection? 

Yes, all participants were informed of the data collection purpose, process and the intended use of the data. They were sent a Participant Information Sheet and signed Consent Form prior to the collection session. All sessions started with an introduction where instructions on data collection, health and safety and confirmation of the Consent Form were discussed.

(j) Did the individuals in question consent to the collection and use of their data?

Yes, all participants were sent a Participant Information Sheet and Consent Form prior to the collection session. We confirmed that they had agreed to and signed the Consent form regarding the collection and use of their data before proceeding with any data collection. Details can be found in Section \ref{sup4}.

(k) If consent was obtained, were the consenting individuals provided with a mechanism to revoke their consent in the future or for certain uses? 

Yes. The Participant Information Sheet and Consent Form addressed how they can request to withdraw and remove their data from the project and how the data will be used.

(l) Has an analysis of the potential impact of the dataset and its use on data subjects (e.g., a data protection impact analysis) been conducted? 

No, all data have been processed to be made de-identifiable and all annotations are on objective world states. The potential impact of the dataset and its use on data subjects were addressed in the Ethics Approval, Participant Information Sheet and Consent Form. Details can be found in Section \ref{sup4}.

(m) Any other comments? 

None. 

4. Preprocessing, Cleaning and Labeling 

(a) Was any preprocessing/cleaning/labeling of the data done (e.g., discretization or bucketing, tokenization, part-of-speech tagging, SIFT feature extraction, removal of instances, processing of missing values)? 

Yes, we have cleaned the videos by blurring participants’ faces. We have also extracted I3D features from the video for action segmentation benchmarking.

(b) Was the "raw" data saved in addition to the preprocessed/cleaned/labeled data (e.g., to support unanticipated future uses)? 

No, we only provide the cleaned videos (participants’ faces being blurred) to the public due to the ethics issues. 

(c) Is the software used to preprocess/clean/label the instances available? 

Yes, we used CVAT to draw bounding boxes. Details can be found in Section \ref{sup2.3}.

(d) Any other comments? 

None. 

5. Uses

(a) Has the dataset been used for any tasks already? 

No, the dataset is newly proposed by us. 

(b) Is there a repository that links to any or all papers or systems that use the dataset? 

Yes, we provide the link to all related information on our website. 

(c) What (other) tasks could the dataset be used for? 

The dataset can also be used for Compositional Action Recognition, Human-Object Interaction Detection, and Visual Question Answering. 

(d) Is there anything about the composition of the dataset or the way it was collected and preprocessed/cleaned/labeled that might impact future uses? 

We granulated the assembly action annotation into subject, action verb, manipulated object, target object and tool. We believe the fine-grained and compositional annotations can be used for more detailed and precise descriptions of the assembly process, and the descriptions can serve various real-world industrial applications, such as robot learning, human robot collaboration, and quality assurance.

(e) Are there tasks for which the dataset should not be used? 

The usage of this dataset should be limited to the scope of assembly activity or task understanding, e.g., action recognition, action segmentation, action anticipation, human-object interaction detection, visual question answering, and the downstream industrial applications, e.g., robot learning, human-robot collaboration, and quality assurance. Any work that violates our Code of Conduct are forbidden. Code of Conduct can be found at our website\footnote{\url{https://iai-hrc.github.io/ha-vid}.}.

(f) Any other comments? 

None. 

6. Distribution 

(a) Will the dataset be distributed to third parties outside of the entity (e.g., company, institution, organization) on behalf of which the dataset was created? 

Yes, the dataset will be made publicly available.

(b) How will the dataset will be distributed (e.g., tarball on website, API, GitHub)? 

The dataset could be accessed on our website. 

(c) When will the dataset be distributed? 

We provide private links for the review process. Then the dataset will be released to the public after the review process. 

(d) Will the dataset be distributed under a copyright or other intellectual property (IP) license, and/or under applicable terms of use (ToU)? 

We release our dataset and benchmark under CC BY-NC 4.0\footnote{\url{https://creativecommons.org/licenses/by-nc/4.0/}.} license. 

(e) Have any third parties imposed IP-based or other restrictions on the data associated with the instances? 

No. 

(f) Do any export controls or other regulatory restrictions apply to the dataset or to individual instances? 

No. 

(g) Any other comments? 

None. 

7. Maintenance 

(a) Who is supporting/hosting/maintaining the dataset? 

Regina Lee and Hao Zheng are maintaining, with continued support from Industrial AI Research Group at The University of Auckland.

(b) How can the owner/curator/manager of the dataset be contacted (e.g., email address)? 

E-mail addresses are at the top of the paper. 

(c) Is there an erratum? 

Currently, no. As errors are encountered, future versions of the dataset may be released and updated on our website. 

(d) Will the dataset be updated (e.g., to correct labeling errors, add new instances, delete instances’)? 
Yes, see 7.(c). 

(e) If the dataset relates to people, are there applicable limits on the retention of the data associated with the instances (e.g., were individuals in question told that their data would be retained for a fixed period of time and then deleted)? 

No. 

(f) Will older versions of the dataset continue to be supported/hosted/maintained? 

Yes, older versions of the dataset and benchmark will be maintained on our website. 

(g) If others want to extend/augment/build on/contribute to the dataset, is there a mechanism for them to do so? 

Yes, errors may be submitted to us through email. 

(h) Any other comments? 

None.


\end{document}